\documentclass[aps,pra,showpacs,notitlepage,twocolumn,superscriptaddress,nofootinbib,preprintnumbers]{revtex4-1}
\usepackage{bbm}
\usepackage{mathrsfs}
\usepackage{epsfig}
\usepackage{soul,color}
\usepackage{graphicx}
\usepackage{amsfonts}
\usepackage{amsthm}
\usepackage[figuresright]{rotating}
\usepackage{amssymb}
\usepackage{amsmath}
\usepackage{dcolumn}
\usepackage{physics}
\usepackage{float}
\usepackage{bm}
\usepackage{verbatim}
\usepackage{braket}
\usepackage[normalem]{ulem}
\usepackage[ruled,vlined,linesnumbered]{algorithm2e}
\usepackage{setspace}
\usepackage[colorlinks,linkcolor=blue,anchorcolor=blue,citecolor=blue,urlcolor=blue]{hyperref}
\usepackage{stackengine}
\usepackage{qcircuit}
\usepackage[capitalise]{cleveref}

\graphicspath{{./figures/}}

\DeclareMathOperator*{\argmax}{arg\,max}
\newcommand{\enc}[1]{\mathrm{Enc}\left[#1\right]}
\newcommand{\nenc}[1]{\widetilde{\mathrm{Enc}}\left[#1\right]}
\newcommand{\dec}[1]{\mathrm{Dec}\left[#1\right]}
\newcommand{\ndec}[1]{\widetilde{\mathrm{Dec}}\left[#1\right]}
\newcommand{\encAND}[1]{\mathrm{Enc}^\land\left[#1\right]}
\newcommand{\encOR}[1]{\mathrm{Enc}^\lor\left[#1\right]}
\newcommand{\encXOR}[1]{\mathrm{Enc}^\oplus\left[#1\right]}
\newcommand{\encNOT}[1]{\mathrm{Enc}^\neg\left[#1\right]}

\newcommand{\notgate}{\textsc{not}}
\newcommand{\andgate}{\textsc{and}}
\newcommand{\orgate}{\textsc{or}}
\newcommand{\xorgate}{\textsc{xor}}
\newcommand{\nand}{\textsc{nand}}

\begin{document}
\title{Fault-Tolerant Neural Networks from Biological Error Correction Codes}
\author{Alexander Zlokapa}
\affiliation{Center for Theoretical Physics, Massachusetts Institute of Technology, Cambridge, MA 02139}
\affiliation{The NSF AI Institute for Artificial Intelligence and Fundamental Interactions} 
\author{Andrew K. Tan}
\affiliation{Department of Physics, Massachusetts Institute of Technology, Cambridge, MA 02139}
\affiliation{The NSF AI Institute for Artificial Intelligence and Fundamental Interactions} 
\author{John M. Martyn}
\affiliation{Center for Theoretical Physics, Massachusetts Institute of Technology, Cambridge, MA 02139}
\affiliation{The NSF AI Institute for Artificial Intelligence and Fundamental Interactions}
\author{Ila R. Fiete}
\affiliation{McGovern Institute for Brain Research, Department of Brain and Cognitive Sciences, Massachusetts Institute of Technology, Cambridge, MA 02139}
\author{Max Tegmark}
\affiliation{Department of Physics, Massachusetts Institute of Technology, Cambridge, MA 02139} 
\affiliation{The NSF AI Institute for Artificial Intelligence and Fundamental Interactions}
\author{Isaac L. Chuang}
\affiliation{Department of Physics, Massachusetts Institute of Technology, Cambridge, MA 02139}
\affiliation{Department of Electrical Engineering and Computer Science, Massachusetts Institute of Technology, Cambridge, MA 02139}
\affiliation{The NSF AI Institute for Artificial Intelligence and Fundamental Interactions}

\begin{abstract}
It has been an open question in deep learning if fault-tolerant computation is possible: 
can arbitrarily reliable computation be achieved using only unreliable neurons? 
In the grid cells of the mammalian cortex, analog error correction codes have been observed to protect states against neural spiking noise, but their role in information processing is unclear.
Here, we use these biological error correction codes to develop a universal fault-tolerant neural network that achieves reliable computation if the faultiness of each neuron lies below a sharp threshold; remarkably, we find that noisy biological neurons fall below this threshold.
The discovery of a phase transition from faulty to fault-tolerant neural computation suggests a mechanism for reliable computation in the cortex and opens a path towards understanding noisy analog systems relevant to artificial intelligence and neuromorphic computing.
\end{abstract}

\preprint{MIT-CTP/5395}
\maketitle

\section{Introduction}
\label{sec:intro}
\begin{figure}[tbp]
        \centering
        \includegraphics[width=0.9\columnwidth]{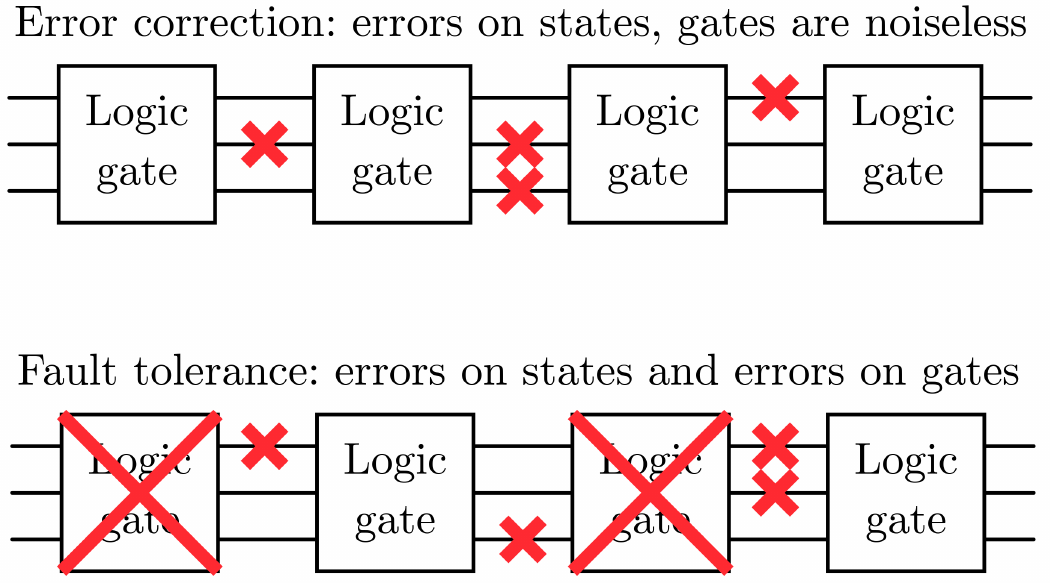}
        \caption{Schematic comparison of error correction and fault-tolerance.
                While error correction uses noiseless gates to correct errors (red crosses), fault-tolerance must use faulty gates to generate reliable output.
                Note that errors on states in the fault-tolerant setup can be rewritten as errors on gates, i.e., faulty wires do not have to be directly considered.
        }
        \label{fig:ec_vs_ft}
    \end{figure}

    Early in the development of computer science, it was unknown if unreliable hardware would make the construction of reliable computers impossible. 
    Whenever a component failed, the resulting error had to be corrected by additional components that were themselves likely to fail. 
    Inspired by ideas from error correction, the notion of \emph{fault-tolerant computation} resolved this issue in standard frameworks of classical and quantum computation~\cite{von-neumann2016probabilistic,pippenger1985on-networks,hajek1991on-the-maximum,evans1998on-the-maximum,evans1999signal,gao2005bifurcations,shor1996fault-tolerant}. 
    In these settings, every computation is evaluated by a sequence of faulty components such as Boolean gates (e.g., \andgate, \orgate, \notgate). 
    If each component's probability of failure falls below a sharp threshold, a strict criterion defining fault-tolerant computation is provably satisfied: computations of any length can be performed with arbitrarily low error.
    It is also worth noting that the distinction between error correction and fault-tolerance is vital here: while error correction uses noiseless gates to correct errors on a state, fault-tolerant computation only has access to faulty gates. We depict this distinction in \cref{fig:ec_vs_ft}.

    In artificial intelligence, it is unresolved~\cite{torres-huitzil2017fault} if neural networks exposed to noise can satisfy an analogous criterion of fault-tolerance.
    That is, taking a noisy neuron as the fundamental component of computation, can any neural network be executed to arbitrary accuracy when the noise strength falls below a threshold?
    A similar question appears in neuroscience, where observations of the mammalian brain have shown that neural \emph{representations} are protected against noise by error correction codes~\cite{hafting2005microstructure,fiete2008what,sreenivasan2011grid}, yet it is unknown if such codes are powerful enough to protect \emph{computations} to achieve arbitrarily small error.
    
    We resolve both open questions in artificial intelligence and neuroscience by demonstrating fault-tolerant neural computation via carefully constructed error correction codes. 
    This success hinges on generalizations of traditional fault-tolerance in Boolean formulas, as well as a modification of a biologically-observed error correction code known as the \emph{grid code}. 
    Beyond the analytic results proven here, we also provide a numerical estimate of the fault-tolerance threshold and show that naturally existing noisy biological neurons lie within the fault-tolerant regime.

    Making the notion of fault-tolerance more precise, we begin by examining von Neumann's original result for fault-tolerant Boolean formulas, which perform universal computation using formulas of Boolean gates~\cite{von-neumann2016probabilistic}.
    In this setting, one considers access to \emph{physical} gates, which are erroneous and fail (i.e., output the incorrect bit) with some fixed probability \(p\).
    In a fault-tolerant construction, each gate in the original error-free formula is replaced by a \emph{logical} gate composed of many physical gates.
    The logical gate is built with error correction such as the repetition code: data is repeated in bundles of three and majority voting determines the outcome.
    Despite the voting itself being performed by faulty physical gates, von Neumann showed via a recursive repetition code that a fault-tolerant Boolean formula can be constructed if the failure probability \(p\) falls below some threshold \(p_0\).
    As Boolean gates suffer discrete errors, we will refer to the fault-tolerance of Boolean formulas as \emph{digital fault-tolerance}, which is formally defined as follows:
        
    \begin{quote}
        \textbf{Digital Fault-Tolerance.} A Boolean formula containing \(N\) (error-free) gates can be simulated with probability of error at most \(\epsilon\) using \(\mathcal{O}(N \text{polylog}(N/\epsilon))\) faulty gates.
        Each gate may fail with probability \(p\) for \(p < p_0\), where \(p_0\) is independent of \(N\) and independent of the noiseless formula depth.
    \end{quote}
    In general, the value of \(p_0\) depends on the model of computation under study~\cite{pippenger1985on-networks,hajek1991on-the-maximum,evans1998on-the-maximum,evans1999signal,gao2005bifurcations}; for example, Ref.~\cite{evans1998on-the-maximum} demonstrated a noise threshold for reliable computation of \(p_0 = (3 - \sqrt{7}) / 4 \approx 0.09\) for Boolean formulas constructed from 2-input \nand~formulas, which are sufficient for universal  computation.

    The digital setting of traditional fault-tolerance strongly contrasts the \emph{analog} computation paradigm of neuroscience and machine learning, where neurons operate using continuous rather than discrete values.
    Here, we consider two biologically-motivated sources of error.
    The first is (1) synaptic failure, where a connection between neurons is dropped~\cite{stevens1994changes, hessler1993the-probability}; this may be modelled by having the neuron output \(0\) with some fixed probability \(p\).
    This is in essence a discrete error (the connection is either present or it is not)
    and may be satisfactorily treated by an extension of von Neumann's construction.
    The second source of error is (2) analog noise afflicting the output of a neuron~\cite{softky1993the-highly}; this may be modelled as additive Gaussian noise with standard deviation \(\sigma\).
    This second type of error is more difficult to correct and will require specialized treatment via the grid code mentioned above.

    To formalize the analog setting of computation, we adopt the framework of artificial neural networks~\cite{mcculloch1943a-logical}, which are universal approximators of continuous functions~\cite{hornik1989multilayer} and have experienced wide success in applications resembling cognitive tasks~\cite{lecun2015deep}.
    The resilience of artificial neural networks to errors has been limited primarily to demonstrations of robustness to weight perturbations or other noise, and hardware fault-tolerance in neuromorphic computing~\cite{neti1992maximally,mhamdi2017when,liu2019a-fault-tolerant,sequin1990fault,neti1992maximally,ruckert1989fault-tolerance}, without considering biologically-motivated noise nor addressing the formal notion of fault-tolerance analogous to digital fault-tolerance defined above.

    We will ultimately prove the following result by using grid-code-based error-correcting mechanisms to achieve fault-tolerant neural computation:
    \begin{quote}
        \textbf{Neural Network Fault-Tolerance.} A Boolean formula of \(N\) (error-free) gates can be simulated by a neural network with probability of error at most \(\epsilon\) using only faulty neurons.
        Each synapse entering a neuron fails with probability \(p\); the output of each neuron is subject to additive Gaussian noise with mean zero and standard deviation \(\sigma\); a neuron admits at most a fixed number of synapses.
        There exist nonzero thresholds \(p_0\) and \(\sigma_0\) such that if \(p < p_0\) and \(\sigma < \sigma_0\), simulating the formula requires \(\mathcal{O}(N \text{polylog}(N/\epsilon))\) faulty neurons.
    \end{quote}
    In the spirit of previous fault-tolerance results \cite{von-neumann1956probabilistic,evans1998on-the-maximum,evans2003on-the-maximum}, the core of our proof is the construction of a logical neuron from a configuration of noisy physical neurons.
    
    An outline of this work is as follows. In \cref{sec:digital-ft}, we first provide a brief review of digital fault-tolerance and then demonstrate how this construction may be adapted to design neural networks that are robust against synaptic failure.
    This is followed by the design of a neuron that is robust to additive Gaussian noise by encoding data in the grid code in \cref{sec:logical-neuron}; here we also demonstrate how error correction and computation may be achieved using a noisy neural network.
    We then showcase our fault-tolerant construction by designing a reliable circuit using our logical neuron subject to both modes of noise in \cref{sec:demonstration-of-computation}, thereby proving our statement of neural network fault-tolerance.
    Finally, we provide some concluding remarks including a discussion of the biological plausibility of our assumptions in \cref{sec:conclusion}.

\section{Fault-tolerance against digital errors}
\label{sec:digital-ft}

First, we provide a review of concatenated fault-tolerance results in digital circuits (\cref{sec:ft-circuits}).
This is followed by a demonstration of an analogous technique for constructing neural networks that are robust against synaptic failure (\cref{sec:FTSynapticFailure}).

\subsection{Fault-tolerant Boolean circuits}
\label{sec:ft-circuits}

    \begin{figure*}[htbp]
        \centering
        \includegraphics[width=0.98\linewidth]{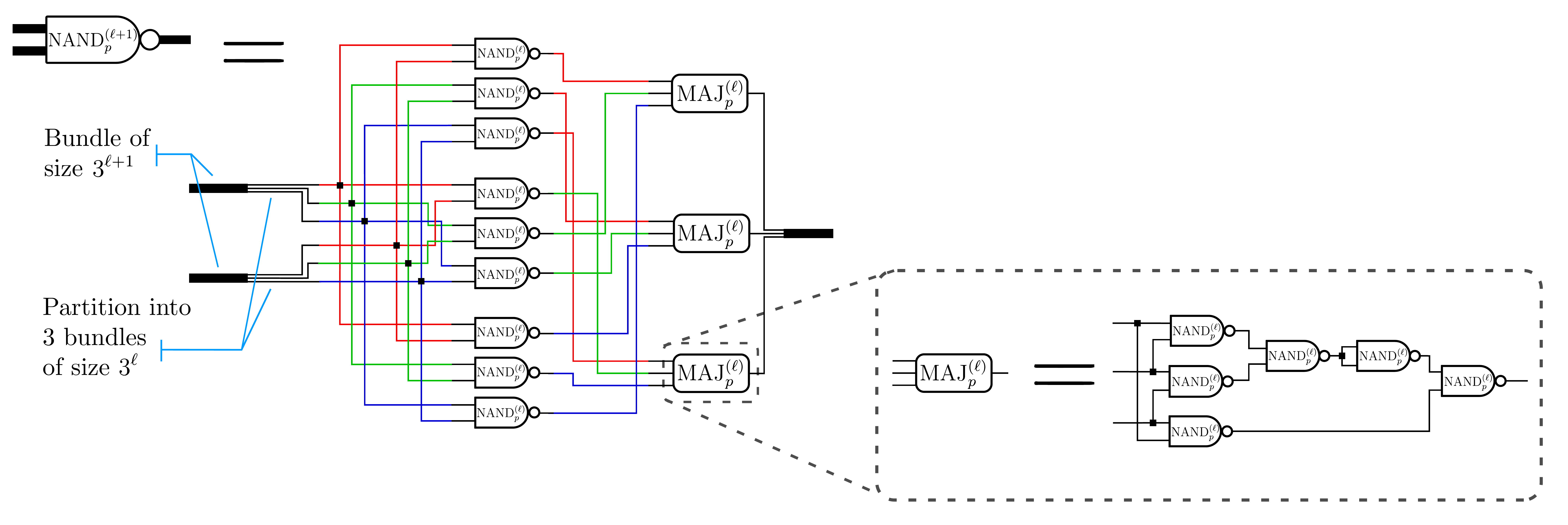}
        \caption{The recursive concatenation scheme, based on a ternary repetition code, used to construct a logical \nand~gate at concatenation level \(\ell+1\) (denoted \nand\(^{(\ell+1)}_p\)) from logical \nand~gates at concatenation level \(\ell\), with the base case \(\nand^{(0)}_p = \nand_p\).
        The gates denoted \(\text{MAJ}^{(\ell)}_p\) indicate a majority voting operation built from \(\nand^{(\ell)}_p\) gates, whose explicit construction is illustrated in the inset. 
        }
        \label{fig:NAND_Concatenate}
    \end{figure*}

    The original construction of a fault-tolerant Boolean gate was initially proposed in Ref.~\cite{von-neumann2016probabilistic} and more rigorously discussed in Ref.~\cite{winograd1963reliable}.
    We begin by presenting an adaptation of this construction via a recursive concatenation of repetition codes.
    To best explain this scheme, let us consider a Boolean gate \(B\), with associated function \(B(x)\) that accepts as input a string of bits \(x\) and outputs a single bit (for instance, \( B(x_0, x_1) = \nand(x_0, x_1) = \neg (x_0 \vee x_1)\)). Let us also consider its faulty counterpart \(B_p\) that fails (i.e., outputs the incorrect bit) with probability \(p\).
    We would like to construct a fault-tolerant version of \(B_p\) whose error can be decreased arbitrarily for \(p<p_0\) for some threhsold $p_0$.

    This is achieved by devising a recursive concatenation scheme wherein a \emph{logical} \(B\) gate is constructed from \emph{physical} \(B\) gates, these being the faulty \(B_p\) gates.
    In particular, a logical \(B\) gate at concatenation level-\(\ell\), which we denote by \(B^{(\ell)}_p\), is recursively defined by a mapping of logical \(B\) gates at concatenation level \(\ell-1\) (i.e., \(B^{(\ell - 1)}_p\)), with the base case \(B^{(0)}_p = B_p\).
    In this mapping, \(B^{(\ell)}_p\) is defined as a repetition code acting on multiple outputs of \(B^{(\ell-1)}_p\), such that the error suffered by \(B^{(\ell)}_p\) is less than that of \(B^{(\ell-1)}_p\) for \(p < p_0\). Thus, increasing $\ell$ decreases the error arbitrarily. 

    In his seminal work on fault-tolerance~\cite{von-neumann2016probabilistic}, von Neumann employed a ternary repetition code, in which a logical bit is encoded as a bundle of physical bits.
    At concatenation level \(\ell\), each bundle consists of \(3^\ell\) physical bits, and its corresponding logical bit may be decoded as the majority of its physical bits.
    For instance, the bundle \(110\) encodes the logical bit \(1\) at concatenation level \(\ell=1\).
    In this manner, the inputs and outputs to \(B^{(\ell)}_p(x)\) are bundles of size \(3^\ell\), and the output is correct if its physical bits decode to the correct logical bit.
     
    The recursive mapping from \(B^{(\ell)}_p\) to \(B^{(\ell+1)}_p\) is defined by this ternary repetition code: the inputs to \(B^{(\ell+1)}_p\) are each linearly partitioned into three smaller bundles, which are then copied and sent through nine \(B^{(\ell)}_p\) gates in parallel to generate nine independent outputs.
    To correct errors in these nine outputs, they are then split into three groups of threes, each of which is passed through a (faulty) majority voting gate, and the three resulting outputs are recombined to represent the final output of \(B_p^{(\ell+1)}\).
    The majority voting gate is constructed from \(B^{(\ell)}_p\) gates, and hence is also imperfect; 
    its explicit construction depends on the Boolean gate of interest and influences the fault-tolerance threshold.
    In general, the fewer the gates in the majority gate, the larger the threshold.
     
    For clarity, we depict this fault-tolerance construction applied to a \nand~gate in \cref{fig:NAND_Concatenate}.
    The specific arrangement of the wires fed into the majority voting gates is chosen is to prevent error propagation and produce a nonzero threshold.
    Not all arrangements will yield a nonzero threshold in the limit \(\ell \rightarrow \infty\); von Neumann's original presentation even suggests randomly permuting these wires.
    Numerics indicate that this particular construction produces a threshold $p_0 \approx 2.36 \% $.
    
    As the \nand~gate is universal for Boolean computation, this construction enables arbitrarily accurate computation of any Boolean function from faulty \nand~gates if the failure probability \(p\) lies below the threshold \(p_0\). 
    Moreover, for \(p<p_0\), the logical error suffered decreases doubly-exponentially with increasing \(\ell\), while the circuit size grows only exponentially with \(\ell\). Hence, achieving a desired error \(\epsilon\) requires overhead \(\mathcal{O}(\text{polylog}(1/\epsilon))\) by the usual arguments for concatenation codes (see e.g. the fault-tolerance threshold theorem of Ref.~\cite{nielsen2010quantum}). 
    
    For a circuit of \(N\) gates, an overall error of \(\epsilon\) could be achieved by demanding individual gate errors \(\epsilon/N\) as per the union bound. Inserting this desired error rate into the above polylogarithmic overhead, we find a total gate count \(\mathcal{O}(N \text{polylog}(N/\epsilon))\), in accordance with the digital fault-tolerance theorem discussed in \cref{sec:intro}.

\subsection{Fault-tolerant neural networks for synaptic failure}
\label{sec:FTSynapticFailure}
    The above fault tolerant construction may be adapted to devise a fault-tolerant neural network that is robust against synaptic failure, as this is in essence a discrete error.
    To illustrate this, let us consider a neural network constructed from rectified linear unit (ReLU) activation functions, where ReLU\((x) = \text{max}(0,x)\) on real inputs \(x\).
    In this case, synaptic failure may be modeled by replacing each ReLU with a faulty ReLU that fails with probability \(p\), i.e.,
    \begin{equation}
        \text{ReLU}_p (x) := 
        \begin{cases}
            \text{ReLU} (x) & \text{with probability } 1-p \\
            0 & \text{with probability } p.
        \end{cases}
    \end{equation}
    Like von Neumann's error model for Boolean gates, the output of this faulty ReLU is incorrect with some probability, and thus its errors may be corrected by employing a concatenated repetition code.
     
\begin{figure}
        \centering
        \includegraphics[width=0.98\columnwidth]{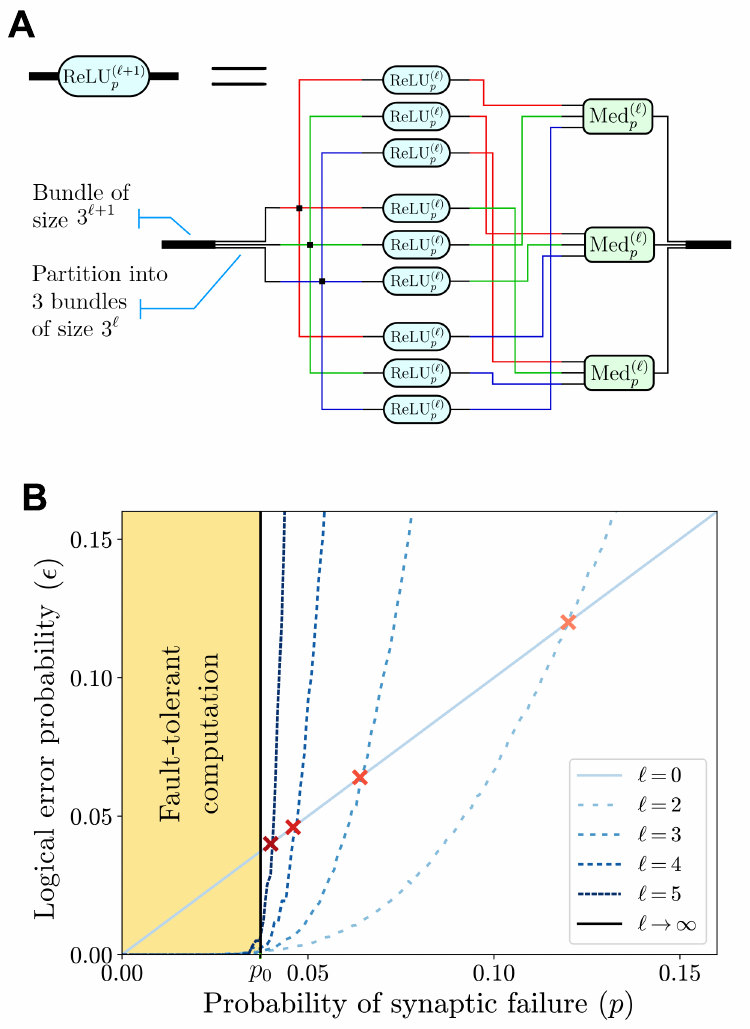}
        \caption{(\textbf{A}) The recursive concatenation scheme of digital fault-tolerance is extended to construct a logical ReLU at concatenation level \(\ell+1\) from logical ReLUs at level \(\ell\).
        Note how the fault-tolerant ReLU is a generalization of the fault tolerant NAND gate in \cref{fig:NAND_Concatenate}.
        The gates denoted \(\text{Med}^{(\ell)}_p\) indicate a median operation that is composed of ReLU\(^{(\ell)}_p\)'s and used to correct errors; its explicit construction is presented in \cref{eq:Median}. 
        This construction ultimately generates a logical neuron for a fault-tolerant neural network in the presence of synaptic failure.
        (\textbf{B}) The logical error probability of ReLU\(^{(\ell)}_p(x)\) on random inputs \(x \in [-1,1]\) as determined by numerical simulation.
        The pseudothresholds (red crosses) occur when the error probability intersects that of \(\ell=0\); they converge exponentially to the threshold \(p_0 \approx 3.72\%\) (vertical black line) with increasing \(\ell\).
    }
        \label{fig:FT_Relu_pseudo}
    \end{figure}

    The aim is to construct a fault-tolerant ReLU activation function, which is equivalent to a fault-tolerant neuron.
    We will employ a concatenated ternary repetition code analogous to that presented above, replacing the logical Boolean gates with logical ReLU's.
    However, there is one important distinction in our construction: as inputs and outputs are now analog instead of binary, we will interpret the logical value carried by a bundle as the \emph{median} of its values rather than the majority.
    Accordingly, the majority voting gate in the original repetition code is replaced by a median gate, which will appropriately correct errors that occur in a bundle.
    With this modification noted, we illustrate the complete recursive scheme in \cref{fig:FT_Relu_pseudo}a; here, it is shown how to construct a logical ReLU at concatenation level \(\ell+1\) (denoted \(\text{ReLU}_p^{(\ell+1)}\)) from logical ReLU's at concatenation level \(\ell\) (denoted \(\text{ReLU}_p^{(\ell)}\)), with the usual base case \(\text{ReLU}_p^{(0)} = \text{ReLU}_p\). 

    What remains is to construct the median operation out of ReLU's.
    At concatenation level \(\ell\), we are interested in computing the median of three bundles, each of size \(3^\ell\).
    Denoting this quantity as \(m=\text{Med}^{(\ell)}_{p}(a,b,c)\), where $a,b,c$ represent each bundle, it may be computed with the following network of depth three:
    \begin{equation}\label{eq:Median}
    \begin{split}
        &x = \text{ReLU}^{(\ell)}_p(a - b) \\
        &y = \text{ReLU}^{(\ell)}_p(-a + c + x) \\
        &z = \text{ReLU}^{(\ell)}_p(b - c + x) \\
        &m = \text{ReLU}^{(\ell)}_p(a + b - c + y - z).
    \end{split}
    \end{equation}
    While the final ReLU is not strictly necessary for the computation of the median, it is included to prevent error propagation and achieve fault-tolerance.
    As a result, this median works only on positive inputs, but this is admissible as the output of ReLU\(^\ell_p\) (which is input into the median) is necessarily non-negative.
    We also note that expressing this median construction as a neural network requires skip connections to perform its computation.

    We visualize the performance of this fault-tolerance construction in \cref{fig:FT_Relu_pseudo} by plotting the \emph{pseudothresholds}: where the error probability at concatenation level \(\ell\) intersects that of \(\ell=0\). Plotting these for increasing levels of concatenation indicates a convergence to the threshold \(p_0 \approx 3.72\%\).
    Therefore, this construction ultimately produces a fault-tolerant ReLU neuron, protected against synaptic failure for \(p < p_0 \).
    And by an argument analogous to the digital fault-tolerance of Boolean circuits, achieving a desired error \(\epsilon\) requires overhead \(\mathcal{O}(\text{polylog}(1/\epsilon))\). Using the argument presented at the end of~\cref{sec:ft-circuits}, this translates to an overhead \(\mathcal{O}(N \text{polylog}(N/\epsilon))\) for a circuit of $N$ gates, thus achieving neural-network fault-tolerance (excluding Gaussian noise) as presented in \cref{sec:intro}.

\section{Fault-tolerance against analog errors}
\label{sec:logical-neuron}

    While a simple adaptation of fault-tolerant constructions on noisy Boolean circuits yields a neural network that is robust to synaptic failure, the treatment of additive Gaussian noise proves more difficult.
    In particular, the repetition-based scheme of von Neumann fails for Gaussian noise with nonzero standard deviation \(\sigma\): unlike the exponential suppression found for digital errors, repeating \(N\) neurons in the presence of analog noise only reduces analog noise to \(\sigma/\sqrt{N}\). Hence the requisite circuit size scales as $O(1/\epsilon^2)$, which does not achieve the $O(\text{polylog}(1/\epsilon))$ performance desired by the neural network fault-tolerance theorem.

    Instead, we turn to an analog error correction code: the grid code.
    Unlike the repetition code, the grid code achieves exponentially small error at asymptotically finite information rates, saturating the Shannon bound~\cite{goblick1965theoretical} and allowing effective error correction against Gaussian neural spiking noise~\cite{sreenivasan2011grid} (see \cref{sec:AnalogRepetitionScaling} for a more detailed discussion).

    We start with a brief overview of the grid code and its properties in \cref{sec:grid-code-overview}.
    Next, we detail the construction of an error correcting procedure using noisy neurons in \cref{sec:nn-error-correction}.
    Finally, we describe in \cref{sec:ft-gaussian-nn} how the logical signal may be manipulated in a manner that allows for universal approximation and analyze its error threshold assuming a distribution of logical neural weights.

    \subsection{Overview of the grid code}
    \label{sec:grid-code-overview}
    We first provide a brief, self-contained exposition of the original grid code results of Refs. \cite{hafting2005microstructure,fiete2008what,sreenivasan2011grid}.
    These works study the entorhinal cortex in mammals and show that lattice neural firing patterns may correspond to a special encoding of the mammal's position (in 2D space), known as the \emph{grid code}. 
    In the grid code, a particular coordinate (say \(x\) or \(y\) in 2D space) takes values from a discrete set \(\{ x_k \}\) of \(S\) possible values that lie within a fixed interval \([0, X)\).

\begin{figure}
        \centering
        \includegraphics[width=0.98\columnwidth]{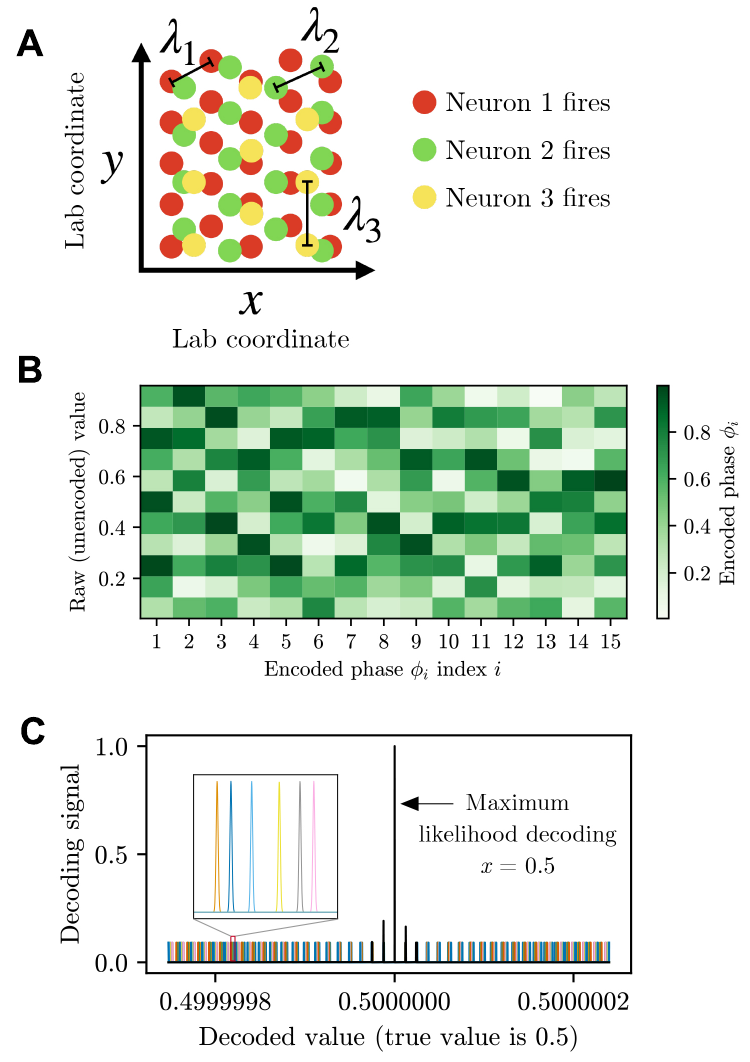}
        \caption{(\textbf{A}) Biological setting of the grid code.
        Neuron firings form a hexagonal lattice with different spacings \(\lambda_j\), with lattice sites corresponding to physical locations of an animal in the lab.
        (\textbf{B}) Example encoding performed by the grid code over \(M=15\) moduli \(\{\lambda_1, \dots, \lambda_{15}\}\). Observe that these phases are well-approximated as being drawn uniformly at random, in accordance with the formalism of the grid code.
        (\textbf{C}) Example decoding of phases representing \(x=0.5\).
        The possible decodings allowed by a given phase (indicated by a unique color for each \(\lambda_j\)) are periodic.
        Each decoded phase is subject to Gaussian noise (inset).
        Since the phases constructively add at the true decoded value, maximum likelihood estimation selects the value with the highest signal.
        }
        \label{fig:grid}
    \end{figure}

    The encoding of each possible value \(x_k\) is modeled by a set of phases
    \begin{equation}
    \label{eq:enc}
        \enc{x_k} := \left\{\frac{e(x_k)}{\lambda_j} \bmod 1\right\}_{j=1}^M,
    \end{equation}
    which is defined over \(M\) relatively prime integers \(\{ \lambda_j\}_{j=1}^M\), referred to as \emph{moduli}~\cite{hafting2005microstructure,fiete2008what}, and a function \(e(x)\) referred to as the \emph{encoding function}.
    The choice of relatively prime moduli ensures, by consequence of the Chinese Remainder Theorem, that all \(x \in [0, \prod_{j=1}^M \lambda_j)\) are encoded into distinct codewords.
    Restricting our domain as above, with \(X \ll \prod_{j=1}^M \lambda_j\), allows the remaining phase space to be used for error correction.
    Moreover, in the original grid code, the encoding function \(e(x)\) is chosen to be the identity.
    Here, we will instead perform neural network computations by selecting \(e(x)\) to implement an activation function; we will let \(e(x)\) be an arbitrary function for now, and specify it later. 
    In general, we denote the vector of \(M\) phases produced by the encoder as \(\boldsymbol \phi := \enc{x_k} = \{\phi_j\}\).
    An example of a firing pattern of the grid code, as well as its moduli, is illustrated in \cref{fig:grid}a.

    To maintain the favorable error-correcting properties of the grid code, the \(x_k\)'s are chosen to satisfy \(x_k \ll X\), and the minimum spacing between codewords \(\Delta x := \min_{k \ne j} |x_k - x_j|\) is chosen such that \(\max_j \lambda_j \ll \Delta x\).
    More generally, when the encoding function \(e(x)\) is not the identity function, the same condition must be upheld for \(\Delta x\) given by \(\min_{i \ne j} |e(x_i) - e(x_j)|\) such that \(e(x_i) \neq e(x_j)\).

    In the limit \(\max_j \lambda_j \ll X\) for \(X \ll \prod_{j=1}^M \lambda_j\), the codeword \(\boldsymbol{\phi}\) encoding a randomly sampled \(x_{k^*} \in \{ x_k \}\) is well-approximated as being drawn from a uniform distribution (\(\phi_j \sim \mathcal{U}(0, 1)\))~\cite{sreenivasan2011grid}.
    We visualize this fact in \cref{fig:grid}b by plotting the phases of an example grid code.
    This property provides a sensitive encoding that changes significantly if the input is slightly perturbed.
    Since each codeword consists of a vector of phases \(\{ \phi_j \}\) with the period of each \(\phi_j\) determined by \(\lambda_j\), decoding corresponds to the constructive interference of summed phases to yield the correct decoded position, as depicted in \cref{fig:grid}c.

    An ideal decoder \(\dec{\boldsymbol{\phi}}\) would perform maximum likelihood estimation (MLE) to recover the most probable value \(x_{k^*}\) given a codeword \(\boldsymbol\phi\).
    For ease of presentation, we modify the original biologically inspired neural decoder that approximates MLE~\cite{sreenivasan2011grid} to a simpler but functionally equivalent form; this form will be more easily implemented by a neural network later in this work.
    Given phases \(\boldsymbol\phi = \{\phi_j\}\), we will recover the true position \(x_{k^*}\) by the MLE decoder
    \begin{equation}
    \label{eq:dec}
        \dec{\boldsymbol \phi} := \argmax_{x_k} \sum_{j=1}^M \cos\left[2\pi\left(\frac{x_k}{\lambda_j} - \phi_j\right)\right].
    \end{equation}

    To see that this procedure is indeed performing maximum likelihood estimation, observe that if \(x_{k^*}\) is known to belong to a discrete set of values \(\{x_k\}\), then the estimated decoding \(\hat x\) is given by maximizing the conditional probability
    \begin{equation}
      \hat x = \argmax_{x_k} \operatorname{Pr}(\boldsymbol\phi | x_k).
    \end{equation}
    Assuming that the encoding \(\enc{x_k}\) is distributed in the codespace according to a spherical Gaussian with variance \(s^2\), the likelihood function is a wrapped normal distribution
    \begin{equation}
        \operatorname{Pr}(\boldsymbol\phi | x_k) \propto \prod_{j=1}^M \exp\left(-\frac{1}{2 s^2} \|\enc{x_k}_j - \phi_j\|^2 \right),
    \end{equation}
    where \(\|\phi\| := \min\{|\phi|, 1 - |\phi|\}\) denotes the distance between phases.
    In the limit of \(s \ll 1\) and taking $e(x_k) = x_k$, the likelihood function is well approximated by the more tractable circular normal function
    \begin{equation}
        \operatorname{Pr}(\boldsymbol\phi | x_k) \propto \prod_{j=1}^M \exp\left(\frac{1}{2 \pi s^2} \cos\left[2 \pi \left( \frac{x_k}{\lambda_j} - \phi_j \right) \right] \right).
        \label{eq:approximate-likelihood}
    \end{equation}
    Comparing \cref{eq:approximate-likelihood} to \cref{eq:dec}, we see that the decoding scheme of \cref{eq:dec} is indeed maximizing the likelihood.

    Lastly, to more intuitively understand the grid code, note that because the \(M\) phases \(\phi_j\) fall between 0 and 1, the coding space is the unit hypercube \([0, 1]^M\); due to the unit modulo, the coding space satisfies periodic boundary conditions and thus corresponds to the \(M\)-torus.
    The coding line \([0, X)\) is thus a set of parallel line segments in the hypercube.
    In general, error correction codes may be described as a hypersphere packing problem: each codeword corresponds to an origin of a sphere in a high-dimensional space, and errors that fall within the radius of the sphere are correctable to the true codeword.
    Here, the grid code is a hypersphere packing problem in the \(M - 1\) dimensional hyperplane perpendicular to the coding line segments.
    Under this formalism, we arrive at a scaling of the minimum distance between line segments with the number of phases for fixed \(X\) of \(d_{\textrm{min}} = \Theta(\sqrt{M})\) \cite{sreenivasan2011grid}, denoting an asymptotic bound on \(d_{\textrm{min}}\) from both above and below. 
    Our choice of \(\lambda \ll \Delta x\) ensures that each \(\enc{x_i}\) lies within a different line segment and is therefore also separated by at least \(d_{\textrm{min}}\), and consequently any perturbation in the phase space less than \(d_{\textrm{min}} / 2\) is correctable using the maximum likelihood decoder.

    \subsection{The fault-tolerant logical neuron}
    \label{sec:nn-error-correction}    
    Let us now use the grid code to present and analyze the construction of a fault-tolerant neuron. 
    We assume an error model where Gaussian noise \(\xi \sim \mathcal{N}(0, \sigma)\) is added to the output of each neuron, representing the noise associated with neural spikes in a biological setting. 
    We note that we do not account for synaptic failures at this stage, as the grid code is only tailored to analog noise; later in \cref{sec:gaussian-and-synaptic-reliability}, we address both additive Gaussian noise and synaptic failure.

    Focusing on a single neuron in a larger neural network, we take the number of neurons connected from the previous layer to be \(m_0\). 
    As in the presentation of the grid code, each neuron carries a value that is guaranteed to belong to a discrete set of \(S\) values, which we parameterize here as \(\{x_k = k\Delta x\}\) for \(k=0, \dots, S-1\), such that \((S-1)\Delta x < X\) for some \(X\).
    The \(M\) relatively prime moduli must satisfy \(\lambda_j \ll X \ll \prod_{j=1}^M \lambda_j\), and thus the codewords are uniformly distributed for random \(x_{k^*}\).

    We however introduce the following modification to the underlying grid code. 
    While the typical grid code assumes a range of values \( x_k \in [0, X)\), here we will take advantage of the periodicity of the grid code due to the periodicity of the phases (as these are evaluated modulo 1), and introduce a smaller range of values \([0, X')\) to which the encoding function \(e(x)\) may output.
    That is, the encoding function $e(x)$ is chosen such that encoded values \(e(x_{k})\) exist in a condensed space \([0,X')\) for some \(X'<X\), while fully decoded values \(x_k\) can still exist in the larger space \([0,X)\).
    A vanilla grid code with an identity encoding function \(e(x_k) = x_k\) has \(X'=X\); here, we will select \(e(x_k)\) to be a non-identity function with \(X'<X\), which will assist in building logical activation functions and thus performing neural network computation.
    Moreover, to remain consistent with the usual requirement that \(\lambda_j \ll X \ll \prod_{j=1}^M \lambda_j\), we will also demand \(\lambda_j \ll X' \ll \prod_{j=1}^M \lambda_j\).

    Turning now to the construction of our fault-tolerant neuron, we incorporate the traditional principles of fault tolerance: we perform computations in the codespace to protect against errors, and interleave each computation between encoding and decoding steps that correct errors and ensure that computation remains in the protected codespace. 
    In the language of the grid code, this means performing computations on the phases \(\boldsymbol{\phi}\), these computations corresponding to the application of weights and biases followed by an activation function. 
    Note that the result of our decoding step is a one-hot encoding of \(x_k\), rather than \(x_k\) itself, and therefore the decoded signal remains redundantly encoded and protected from noise;
    this is in line with traditional fault-tolerant constructions where the signal is maintained redundantly throughout computation and error correction.
    
\begin{figure}
        \centering
        \includegraphics[width=0.98\linewidth]{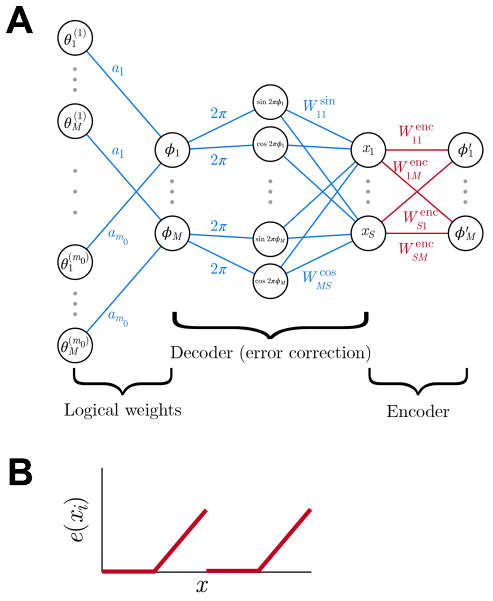}
        \caption{(\textbf{A}) Logical neuron decomposed into physical neurons to achieve fault-tolerance in the presence of analog noise.
        The neuron receives encoded neural outputs from the previous layer and performs a computation with time advancing to the right.
        The logical weights \(a_i\) are applied in the codespace, and the decoder recovers \(x_k\) by performing error correction.
        The encoder (red) performs a logical activation function (e.g., ReLU) using appropriate weights and encodes back to the codespace.
        (\textbf{B}) The logical ReLU encoding function \(e(x_i)\) used in the encoder (\cref{eq:relu}) for \(X' = X/2\). 
        This function is a ReLU repeated over $X/X' = 2$ periods; the construction allows the fault-tolerant neural network to implement the standard ReLU activation function.
        }
        \label{fig:logical_neuron_relu}
    \end{figure}

    The general construction of the logical neuron is presented in \cref{fig:logical_neuron_relu}a. This depicts a logical neuron decomposed into physical neurons, with time advancing to the right. 
    The number of inputs to the physical neurons is unrestricted, and hence this construction has an unbounded fan-in.

    In the illustration, a previous layer of logical neurons passes to the logical neuron a set of encoded phase vectors, which we denote as \(\boldsymbol{\theta}^{(i)} = [\theta^{(i)}_1, ..., \theta^{(i)}_M]\) for the \(M\)-dimensional phase vector of the \(i\)th logical input neuron.
    Assuming that inputs to the network are all encoded in the same grid code, each input phase vector \(\boldsymbol{\theta}^{(i)}\) encodes a quantity that lies in \([0, X')\) in the decoded space.

    The logical neuron itself consists of three stages: (1) the logical weights, (2) the decoder, and (3) the encoder. 
    First, (1) the logical weights correspond to the weights of the error-free neuron that one seeks to apply; we denote these by \(\{a_i\}_{i=1}^{m_0}\) for each of the \(m_0\) (logical) neurons of the previous layer.
    As depicted in the figure, the logical weights are each repeated \(M\) times and then applied to the inputs \(\boldsymbol{\theta}^{(i)}\), mapping directly from grid code phases to grid code phases.
    
    Second, (2) the decoder performs error correction via maximum likelihood estimation (MLE) as described in \cref{sec:grid-code-overview}.
    The key observation is that the structure of the grid code allows MLE to be approximated by a neural network.
    This is achieved using sine and cosine activation functions with appropriately chosen weights, the combination of which implements the MLE decoding scheme of \cref{eq:dec} and also imposes the periodicity of the resulting phase encoding. 
    The particular choice of weights, denoted \(W^{\text{sin}}_{ik}\) and \(W^{\text{cos}}_{ik}\), is explained and justified in the following section.
    Moreover, the decoder does not return to the original \([0, X')\) space; instead, it outputs a value in the larger space \([0, X)\), allowing the application of logical weights to decode to valid values.
    At the end of the decoding step, we are left with one-hot encoding representing the correct value \(x_k \in [0, X)\) with high probability due to the robustness of the maximum likelihood estimate.
    
    Lastly, (3) the encoder serves two roles: it re-encodes back into the codespace, and it also performs the computation via the application of a logical activation function (e.g., ReLU).
    Explicitly, the weights \(W_{ki}^{\text{enc}}\) are chosen such that this stage projects the one-hot representation of some \(x_k\) back to its appropriate codeword. The specific choice of weights also applies the logical activation function through a chosen encoding function \(e(x_k)\), ultimately returning to the space \([0, X')\). The choice of weights \(W_{ki}^{\text{enc}}\) and encoding function $e(x)$ for various activation functions are presented in the following sections.

    \subsection{Neural network implementation of reliable computation}
    \label{sec:ft-gaussian-nn}
    Let us now analyze the performance of the logical neuron in the fault-tolerant setting, where every physical neuron is subjected to noise. We will ultimately derive an analytical expression for the number of physical neurons needed to build a logical neuron with logical error at most \(\epsilon\).

    To streamline our presentation, we begin by looking at the encoder stage.
    Accounting now for the additive Gaussian noise \(\xi \sim \mathcal{N}(0, \sigma)\) suffered by the physical neurons, the encoder of \cref{eq:enc} becomes
    \begin{equation}
        \label{eq:nenc}
        \nenc{x_{k}} = \left\{\tilde{\phi}_j = \frac{e(x_{k})}{\lambda_j} + \xi \bmod 1\right\},
    \end{equation}
    for i.i.d.~\(\xi \sim \mathcal{N}(0, \sigma)\) sampled for each phase $\tilde{\phi}_j$.
    The decoder in the logical neuron uses only two layers (see \cref{fig:logical_neuron_relu}a).
    The first layer multiplies each phase \(\phi_j\) by a weight \(2\pi\); the second layer uses sine and cosine activation functions to compute \(\sin(2\pi \phi_j)\) and \(\cos(2\pi\phi_j)\), and then multiplies them by weights \(W^{\sin}_{jk}\) and \(W^{\cos}_{jk}\), respectively. We select these weights to be \(W^{\sin}_{jk} = \sin\left(2\pi\frac{x_k}{\lambda_j}\right)\) and \(W^{\cos}_{jk} = \cos\left(2\pi\frac{x_k}{\lambda_j}\right)\).
    
    Upon applying a decoding, any error is `reset' if the decoding \(\ndec{\tilde{\boldsymbol{\phi}}}\) is successful, such that the logical neuron will not propagate any additional error into future computations.
    Evaluating all noise contributions, we have
    \begin{align}
    \label{eq:ndec}
        \dec{\tilde{\boldsymbol{\phi}}} &= \argmax_k f(k),\\
        f(k) :&=  \xi + \sum_{j=1}^M \left[f_1(j, k) + f_2(j, k)\right],
    \end{align}
    where
    \begin{align}
        f_1(j, k) &= \sin\left(2\pi\frac{x_k}{\lambda_j}\right)\left[\sin\left(2\pi\tilde \phi_j\right) + \xi\right],\\
        f_2(j, k) &= \cos\left(2\pi\frac{x_k}{\lambda_j}\right)\left[\cos\left(2\pi\tilde \phi_j\right) + \xi\right],
    \end{align}
    and as usual each \(\xi\) is sampled i.i.d.

    Suppose that the correct neuron value corresponds to \(k = k^*\), i.e., the value \(x_{k^*}\) is encoded in the phases.
    For the decoder to identify the correct neuron via a threshold cutoff, we require \(f(k = k^*) > f(k \neq k^*)\) for all \(k\).
    If the mean of the correct neuron is greater than the mean of each incorrect neuron, a threshold will exist to distinguish the correct decoding from incorrect decodings in expectation.
    We use this insight to gain an analytical scaling for the number of physical neurons needed to construct a logical neuron with logical error \(\epsilon\).

    The key observation is that in the noiseless limit, the phases \(\phi_j\) are given by \(\phi_j = x_{k^*}/\lambda_j \bmod 1\).
    At \(k = k^*\), the elements in the sum of \cref{eq:ndec} constructively add as \(f_1(j, k^*) + f_2(j, k^*) \approx 1\) for each \(j\) and thus \(f(k = k^*)\) has a non-zero mean. 
    On the other hand, for all \(k \neq k^*\), the neural network weights are sine or cosine of a uniformly distributed random variable and the terms in the sum destructively interfere leaving \(f(k \ne k^*) \approx 0\) on average.

    To make the scaling argument precise for computation, we need to characterize the noise in \cref{eq:ndec}, which requires assumptions to be made about the statistical properties of the noise and logical weights.
    With this in mind, we make the following assumptions. 
    In order to maintain properties of modular arithmetic, we restrict the logical weights to integer values \(a_i \in \mathbb{Z}\) such that \(\sum_i |a_i| \leq X/X'\). We also assume that the logical weights \(a_i\) are approximately normally distributed from a Gaussian distribution with standard deviation \(\alpha\).
    Additionally, we take both the number of moduli \(M\) and the number of neurons \(m_0\) connected from the previous layer to be much larger than one, allowing application of the central limit theorem.
    
    As an example, we will select ReLU as the logical activation function of the logical neuron; other activation functions may be implemented analogously. A ReLU activation function may be implemented by the encoding function
    \begin{equation}
    \label{eq:relu}
        e(x_i) = \begin{cases}
        0 & (x_i \bmod X') < X'/2\\
        (x_i - X'/2) \bmod X' & (x_i \bmod X') \geq X'/2,
        \end{cases}
    \end{equation}
    which behaves like a periodic ReLU. We depict this encoding function in \cref{fig:logical_neuron_relu}b.
    This function corresponds to choosing weights on the physical neurons \(W^\mathrm{enc}_{ij} = \frac{e(x_i)}{\lambda_j}\bmod 1\).

    The above analysis can now be made explicit to demonstrate the fault-tolerant properties of the logical neuron.
    Looking at the logical weights stage, the \(m_0\) logical neurons from the previous layer connected to the logical neuron are represented by codewords \(\enc{x^{(1)}}, \dots, \enc{x^{(m_0)}}\), i.e., \(\enc{x^{(i)}} = \boldsymbol{\theta}^{(i)}\). Therefore, the application of the logical weights must map from the \(m_0\times M\) neurons in \(\{\enc{x_i}\}\) to a single set of phases \(\{\phi_j\}\) such that our activation function is applied \emph{after} decoding and re-encoding, as per the order of operations in the logical neuron.
    By assigning weights \(W_{ij} = a_i\) from \(\theta_j^{(i)}\) to \(\phi_j\) (as illustrated in the `Logical weights' layer in \cref{fig:logical_neuron_relu}a) with a linear activation function and bias \(-\frac{X'}{\lambda_j} \sum_{i:a_i < 0} a_i\), we obtain the following phases in the absence of noise:
    \begin{equation}
        \label{eq:phij}
        \phi_j = \left(\sum_{i=1}^{m_0} a_i \theta^{(i)}_j\right) - \left(\frac{X'}{\lambda_j}\sum_{i:a_i < 0}^{m_0} a_i\right).
    \end{equation}
    
    We proceed to include noise in the analysis. Let us denote a noisy encoding by phases \(\tilde \theta_j^{(i)}\).
    Each of the \(S\) neurons over the discretized decoded space have noise \(\xi\), and each is multiplied by approximately uniformly distributed weights due to the phases over the moduli.
    Applying the central limit theorem to \(\sum_{i=1}^S u \xi\) for \(u \sim \mathcal{U}(0, 1)\), we find this equivalent to noise with mean zero noise and variance \(\sigma^2\cdot S/3\).
    Adding this noise of the codespace neuron to the noise acquired from the physical neuron, we find \(\tilde \theta_j^{(i)} = \theta^{(i)}_j + \xi + \zeta\) for \(\zeta \sim \mathcal{N}(0, \sigma \sqrt{S/3})\).
     Inserting noise in \cref{eq:phij}, we have
    \begin{equation}
    \begin{aligned}
        \tilde \phi_j &= \left[\sum_{i=1}^{m_0} a_i \tilde\theta^{(i)}_j\right] - \left(\frac{X'}{\lambda_j}\sum_{i:a_i < 0} a_i\right) + \xi\\
        &= \phi_j + \xi + \sum_{i=1}^{m_0} a_i (\xi + \zeta).
    \end{aligned}
    \end{equation}
    Applying the central limit theorem to the last term \(\sum_{i=1}^{m_0} a_i (\xi + \zeta)\), we find that its mean vanishes while its variance is \(\frac{1}{3}S m_0 \alpha^2\sigma^2\) in the large-\(S\) limit (having already applied the central limit theorem to \(S\)).

    We can now formalize the above argument that correct decoding requires \(f(k = k^*) > f(k \neq k^*)\). To simplify notation, we introduce the variable \(\beta := 4\pi^2(1+S m_0\alpha^2/3)\).
    By applying the error correction analysis (\cref{eq:ndec}) to the phases after a step of computation (\cref{eq:phij}) again in large \(M\) regime, we find that the true decoding after application of the logical neuron is distributed as 
    \begin{equation}
    \label{eq:fk0}
    \begin{aligned}
        &f(k = k^*) \sim \mathcal{N}\Bigg(M e^{-\beta \sigma^2/2},\\
        &\qquad \qquad \quad \sqrt{M\left(\frac{1}{2}+\frac{1}{2}e^{-2\beta\sigma^2}-e^{-\beta\sigma^2}+\sigma^2\right) + \sigma^2}\Bigg),
    \end{aligned}
    \end{equation}
    while the incorrect decoding is centered at zero:
    \begin{align}
    \label{eq:fnk0}
        f(k \neq k^*) &\sim \mathcal{N}\left(0, \sqrt{M\left(\frac{1}{2}+\sigma^2\right)+\sigma^2}\right),
    \end{align}
    where both distributions are seen to have standard deviations \(\mathcal{O}(\sqrt{M})\).
    Upper-bounding the maximum element drawn from the distribution of \(f(k \neq k^*)\) out of \(S\) draws using Jensen's inequality and a union bound, we find that
    \begin{equation}
    \begin{aligned}
        f_\mathrm{max}(k \neq k^*) :&= \mathbb{E}[\text{max draw of } f(k \neq k^*)] \\
        &\leq \sqrt{[M(1+2\sigma^2) + 2\sigma^2]\log S}.
    \end{aligned}
    \end{equation}
    Finally, to determine if \(\argmax_k\) returns a value other than \(k^*\), we compute the probability that this exceeds \(f(k = k^*)\):
    \begin{equation}
        \begin{aligned}
            &\operatorname{Pr}[\text{logical neuron fails}] = \operatorname{Pr}[f(k = k^*) < f_\mathrm{max}(k \neq k^*)]\\
            &\ \ \leq \frac{1}{2}\mathrm{erfc} \Bigg[ \frac{e^{-\beta\sigma^2/2}M - \sqrt{[M(1+2\sigma^2) + 2\sigma^2] \log S}}{\sqrt{M\left(1 + e^{-2\beta\sigma^2}-2e^{-\beta\sigma^2}+2\sigma^2\right) + 2\sigma^2}} \Bigg].
        \end{aligned}
    \end{equation}
    This error probability is the logical error, which we seek to upper bound by \(\epsilon\). Expanding in small \(\epsilon\) and taking \(M, \beta \sigma^2 \gg 1\), we find that the number of moduli required to bound the logical error by \(\epsilon\) scales is
    \begin{equation}
        \begin{aligned}
            M(\epsilon) &\approx \log(1/\epsilon)\left[e^{ \beta \sigma^2}(1+2\sigma^2) + e^{- \beta \sigma^2} - 2\right]\\
            &\approx e^{\beta \sigma ^2}(1+2\sigma^2) \log \left(1/\epsilon\right) = \mathcal{O}(e^{\beta \sigma^2} \log(1/\epsilon)),
            \label{eq:m} 
        \end{aligned}
    \end{equation}
    where \(\beta \) is the aforementioned constant independent of the noise or error correction overhead.
    The \(\mathcal{O}(e^{\beta \sigma ^2})\) dependence on \(\sigma\) originates from the constructive interference of the grid code: noisy phases for the true decoding contribute to a neuron with mean activation \(Me^{- \beta \sigma^2/2}\), while the incorrect decoding yields a mean activation of zero.
    Although the noise produces neural activations of variance \(\mathcal{O}(\sigma\sqrt{M})\), there always exists sufficiently large \(M\) to identify the correct decoding.
    
    Note that the number of physical neurons in the logical neuron scales linearly in the number of moduli, as per its structure in \cref{fig:logical_neuron_relu}a).
    Hence, a fault-tolerant neural network can be constructed under the presence of arbitrarily large additive Gaussian noise using \(\mathcal{O}(e^{\beta\sigma^2} \log(1/\epsilon ) )\) physical neurons for a constant \(\beta\). For a network of $N$ neurons, this translates to $O(e^{\beta\sigma^2} N \text{polylog} (N/\epsilon))$ physical neurons as per the argument of~\cref{sec:digital-ft}.
    This result is in agreement with the neural network fault-tolerance theorem of \cref{sec:intro} (excluding synaptic failure, which we address in \cref{sec:gaussian-and-synaptic-reliability}), and also mirrors known results in digital fault-tolerance~\cite{evans2003on-the-maximum}.

\section{Reliable circuits from the fault-tolerant neuron}
\label{sec:demonstration-of-computation}    
    The fault-tolerant neural network presented above is a universal approximator of continuous functions due to the use of a ReLU activation function.
    In this section, we demonstrate the flexibility of the fault-tolerant neural network construction by building Boolean circuits from fault-tolerant neural networks and providing evidence of their reliability.
    In \cref{sec:gaussian-reliability}, we numerically verify the predictions of \cref{sec:logical-neuron} by simulating the code size requirement to implement a two-bit Boolean multiplication circuit constructed from neurons subject to additive Gaussian noise.
    In \cref{sec:gaussian-and-synaptic-reliability}, we combine the constructions of \cref{sec:digital-ft} and \cref{sec:logical-neuron} to provide analytic and numerical evidence of the robustness of our fault-tolerant neural network against both additive Gaussian noise and synaptic failure.
    Finally, in \cref{sec:repetition-concat}, we move towards more biological code parameters by studying the more biologically realistic scenario where moduli are encoded redundantly.
    
    \subsection{Reliability in the presence of Gaussian noise}
    \label{sec:gaussian-reliability}
\begin{figure*}
        \centering
        \includegraphics[width=0.98\linewidth]{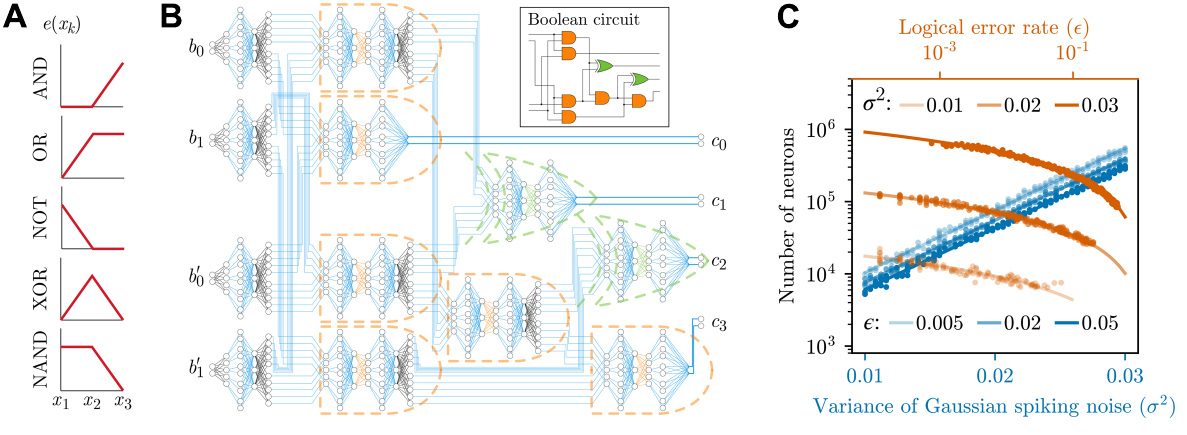}
        \caption{
        (\textbf{A}) Encoding functions \(e(x_k)\) that induce appropriate logical activation functions to implement common Boolean gates.
        All logical weights \(\{ a_i \}\) are set to unity when implementing a Boolean gate.
        (\textbf{B}) Fault-tolerant neural network implementation of the two-bit multiplication circuit.
        In the inset, we illustrate the two-bit multiplication circuit decomposed into six \andgate~gates (orange) and two \xorgate~gates (green).
        In the neural network implementation thereof, two \(2\)-bit binary numbers \(b_0b_1\) and \(b_0'b_1'\) (using the 0 index to denote the least significant bit) are one-hot encoded (as per \cref{eq:boolean_codeword_1,eq:boolean_codeword_2}) as input to the noisy neural network.
        The output \(c_0 c_1 c_2 c_3\) yields the product of the two numbers, and can achieve arbitrarily small error by increasing the number of moduli (\(M=5\) moduli illustrated).
        The grid code corrects Gaussian noise via the decoder (blue lines); neural encoders evaluate \andgate~gates (orange outline) and \xorgate~gates (green outline) to perform computation using appropriate encoder activation functions (shown in \textbf{A}); additional decoders and encoders are used to generate error-corrected copies of neural states (black).
        (\textbf{C}) Numerical simulation of the number of neurons required to perform two-bit multiplication with logical error probability \(\epsilon\) in the presence of Gaussian noise of variance \(\sigma^2\).
        The fit confirms the analytic scaling \(\mathcal{O}(e^{a\sigma^2} N \log(N/\epsilon))\) of \cref{eq:m}.
        }
        \label{fig:booleans}
    \end{figure*}

    Building on the fault-tolerant neural network of \cref{sec:logical-neuron}, a natural extension of this framework to Boolean gates emerges if additional encoding functions are introduced.
    In particular, as computations are done in the encoding step of the fault-tolerant neuron in \cref{sec:nn-error-correction}, special encoding functions can be used to implement \andgate, \orgate, \notgate, \xorgate, and \nand~operations, among other Boolean gates. We illustrate these encoding functions in \cref{fig:booleans}a, whose specific construction we discuss next. Afterwards, we will use these Boolean gate constructions to enable a fault-tolerant neural implementation of a multiplier circuit.
    Because these constructions use the fault-tolerant neuron of \cref{sec:nn-error-correction}, they are robust against Gaussian noise only; we account both Gaussian noise and synaptic failure in \cref{sec:gaussian-and-synaptic-reliability}.

    To formalize this construction, define two logical input bits \(A, B \in \{0, a\}\), interpreting \(0\) as False and \(a\) as True.
    Letting \(\Delta x = a\), the decoder \(\dec{\boldsymbol{\phi}}\) will only decode to the set of variables \(\{x_1 = 0, x_2 = a, x_3 = 2a\}\).
    For notational convenience, we define codeword vectors
    \begin{align}
        \boldsymbol{\phi}_a &:= \left\{\phi_j = \frac{a}{\lambda_j} \bmod 1\right\}, \label{eq:boolean_codeword_1}\\
        \boldsymbol{\phi}_0 &:= \left\{\phi_j = \frac{0}{\lambda_j} \bmod 1\right\} \label{eq:boolean_codeword_2}.
    \end{align}

    Beginning with a \notgate~gate, define the \notgate~encoder \(\encNOT{x_1} = \boldsymbol{\phi}_a\) and \(\encNOT{x_2} = \encNOT{x_3} = \boldsymbol{\phi}_0\).
    As before, this corresponds to a neural network with weights given by the codeword vectors.
    To compute \(\neg A\), we simply compute \(\encNOT{\dec{A}}\), which applies error correction and re-encode into the codespace with a \notgate computation.

    To implement \andgate~and \orgate~gates, we require an additional layer of unity weights, producing the value \(\phi_i = \theta_i^A + \theta_i^B\) for input phases corresponding to bits \(A\) and \(B\).
    Applying the decoder will give either \(0, a\) or \(2a\) based on the cases \((A, B) \in \{(0, 0)\}\), \(\{(a, 0), (0, a)\}\) or \(\{(a, a)\}\) respectively.
    The \andgate~encoder is given by \(\encAND{x_1} = \encAND{x_2} = \boldsymbol{\phi}_0\) and \(\encAND{x_3} = \boldsymbol{\phi}_a\), and the \orgate~encoder is given by \(\encOR{x_1} = \boldsymbol{\phi}_0\) and \(\encOR{x_2} = \encOR{x_3} = \boldsymbol{\phi}_a\).
    
    Likewise the \xorgate~encoder is given by \(\encXOR{x_1} = \encXOR{x_3} = \boldsymbol{\phi}_0\) and \(\encXOR{x_2} = \boldsymbol{\phi}_a\), and the \nand~encoder by \(\text{Enc}^{\bar{\wedge}} [x_1] = \text{Enc}^{\bar{\wedge}} [x_2] = \boldsymbol{\phi}_a\) and \(\text{Enc}^{\bar{\wedge}} [x_3] = \boldsymbol{\phi}_0\). 
    These Boolean gates furnish a universal gate set, from which arbitrary Boolean circuits, and thus arbitrary computations, may be achieved in a fault-tolerant manner. 
    As per the results of \cref{sec:ft-gaussian-nn}, a fault-tolerant neural network assembled of these \emph{neural-Boolean} gates satisfies the neural network fault-tolerance theorem (excluding synaptic failure) with \(\mathcal{O}\left(e^{\mathcal{O}(\sigma^2)} \log(1/\epsilon) \right) \) physical neurons.
    
    As an application of these neural-Boolean gates, we use them to implement a fault-tolerant two-bit multiplier. In this construction, the individual Boolean gates of the two-bit multiplier circuit are replaced with their corresponding neural-Boolean gates.
    We depict this circuit in \cref{fig:booleans}b.
    Here, the neural network takes in two 2-bit binary numbers \(b_0 b_1\) and \(b'_0 b'_1\) and outputs their product, suffering an error that can be decreased arbitrarily error by increasing the number of moduli.
    For this neural two-bit multiplier, we numerically estimate the circuit size required to achieve a logical error rate \(\epsilon\) with respect to the Gaussian noise strength \(\sigma^2\).
    The results are shown in \cref{fig:booleans}c and are in good agreement with the analytic prediction of \cref{eq:m}.

    \subsection{Reliability in the presence of Gaussian noise and synaptic failure}
    \label{sec:gaussian-and-synaptic-reliability}

    Next, we study fault-tolerance with respect to both modes of noise: synaptic failure and additive Gaussian noise.
    Here, we consider a fault-tolerant neural \nand~gate, which simplifies analysis as it is alone sufficient for universal Boolean computation.
    By comparing the \nand~encoding function of \cref{fig:booleans}a and the ReLU encoding function of \cref{fig:logical_neuron_relu}b, we see that the \nand~encoding function is the opposite of the ReLU encoding function.
    Hence, we can transfer over the Gaussian noise analysis of \cref{sec:ft-gaussian-nn} to the setting of the neural \nand~gate, with the modification that we choose \(0\) to correspond to the True state and \(a\) to the False state.
    This makes the ReLU encoding function equivalent to the direct implementation of a \nand~gate.

    Repeating the noisy logical neuron analysis of \cref{eq:fk0}, but now with logical weights \(a_i = 1\) and three decoder neurons, i.e., \(S = 3\), as per the neural \nand~gate construction, we find
    \begin{equation}
    \label{eq:fk0-nand}
    \begin{aligned}
        f_\nand(k^*) \sim \mathcal{N}\Bigg(&M\cdot\frac{e^{-6\pi^2\sigma^2}\erf^6(\sqrt{2}\pi\sigma)}{2^9\pi^3\sigma^6},\\
        &\qquad \qquad \sqrt{M\left(\frac{1}{2}+\sigma^2 - \zeta\right) + \sigma^2}\Bigg),
    \end{aligned}
    \end{equation}
    for
    \begin{equation}
        \zeta = \frac{e^{-12 \pi ^2 \sigma ^2} \operatorname{erf}^{12}\left(\sqrt{2} \pi  \sigma \right)-4 \pi ^3 \sigma ^6 e^{-24 \pi ^2 \sigma ^2} \operatorname{erf}^6\left(2 \sqrt{2} \pi  \sigma \right)}{2^{18} \pi ^6 \sigma ^{12}}.
    \end{equation}
    However, \cref{eq:fnk0} remains unchanged, i.e.
    \begin{align}
    \label{eq:fnk0-nand}
        f_\nand(k \neq k^*) &\sim \mathcal{N}\left(0, \sqrt{M\left(\frac{1}{2}+\sigma^2\right)+\sigma^2}\right).
    \end{align}
    Repeating a similar analysis to estimate \(\operatorname{Pr}[f_\nand(k^*) < f_\nand(k\neq k^*)]\) yields the number of moduli
    \begin{align}
        M(\epsilon) &\approx \frac{2^{18} \pi ^6 e^{12 \pi ^2 \sigma ^2} \sigma ^{12} \left(4 \sigma ^2+1\right) \log \left(\frac{3}{\epsilon }\right)}{\operatorname{erf}^{12}\left(\sqrt{2} \pi  \sigma \right)}\\
        &= \mathcal{O}\left(e^{\beta \sigma^2}\log(1/\epsilon)\right),
    \end{align}
    consistent with the results of \cref{sec:ft-gaussian-nn}.

    To also account for synaptic failure with probability \(p\), we must modify \cref{eq:fk0-nand,eq:fnk0-nand} to include the possibility of this discrete mode of noise.
    While a functional synapse with additive Gaussian noise returns value \(y + \xi\), a synaptic failure returns value \(0\).
    A careful treatment of synaptic failure is provided in \cref{sec:synaptic-failure-details}, the result of which is a new set of distributions \(f_\nand'(k \neq k^*)\) and \(f_\nand'(k = k^*)\) which depend on both the strength of Gaussian errors \(\sigma\) and on the probability of synaptic failures \(p\).
    As with \cref{eq:fk0-nand,eq:fnk0-nand}, \(f_\nand'(k \neq k^*)\) is centered at zero with standard deviation \(\mathcal{O}(\sqrt{M})\), and \(f_\nand'(k = k^*)\) is centered at \(\mathcal{O}(M)\) with standard deviation \(\mathcal{O}(\sqrt{M})\).

    In order to proceed, we must take a more careful treatment of the activation function required for the error correction step of the logical neuron.
    In a biological discussion of the grid code, winner-take-all dynamics are often used to describe the decoding process~\cite{sreenivasan2011grid}, i.e., it is assumed that the only neuron activated is that representing the decoded value with the largest signal, as per maximum likelihood decoding approach discussed in \cref{sec:grid-code-overview}.
    This decoding approach implicitly assumes communication between the decoding neurons, e.g. through an argmax-type non-linearity.
    However, for transparency in the treatment of noise, we demonstrate how a local step activation function, parameterized by a cutoff \(c\), can replace winner-take-all dynamics with a simpler decoder.
    
    Because the separation of means of the correct and incorrect decoding distributions scales as \(\mathcal{O}(M)\) compared to their standard deviations which scale as \(\mathcal{O}(\sqrt{M})\), an appropriate choice of threshold \(c\) is sufficient to distinguish between the two distributions with high probability (for large \(M\)).
    Since there are three decoding neurons, a correct decoding requires the correct neuron sampled from \(f_\nand'(k = k^*)\) to exceed \(c\) and the two incorrect neurons sampled from \(f_\nand'(k\neq k^*)\) to lie below \(c\).
    Evaluating such probabilities is straightforward due to \(f_\nand'(k)\) being normally distributed in both cases.
    The probability that the logical \nand~neuron succeeds is given by \(1 - \epsilon(\sigma, p)\), where \(\epsilon(\sigma, p)\) is an error rate that depends on both the strength of Gaussian errors \(\sigma\) and the probability of synaptic failures \(p\).
    A more detailed analysis of \(\epsilon(\sigma, p)\), including its explicit expression, is provided in \cref{sec:synaptic-failure-details}.
    
    To obtain a fault-tolerance threshold from this quantity, we apply the result of Evans and Pippenger \cite{evans1998on-the-maximum} for fault-tolerant Boolean formulas built from \nand~gates. 
    Evans and Pippenger present a construction for that Boolean formulas built from \nand~gates that achieves fault-tolerance if and only if the \nand~probability of failure is below \(\epsilon_0 = (3-\sqrt 7)/4\).
    We appeal to this bound to prove fault tolerance of neural \nand~gates, which is equivalent to placing the grid code inside the code of Evans and Pippenger.
    While their \nand~construction only considers errors as bit flips -- i.e., an error is triggered if a gate that should return \(0\) returns a \(1\), and vice versa -- errors in the neural \nand~gate are biased.
    This occurs because synaptic failures bias neurons towards zero output; if all neurons fail, the neural \nand~defaults to \(0\).
    However, biased errors are strictly easier to correct than unbiased errors, and thus the threshold of Evans and Pippenger serves as an appropriate lower bound.
    To ensure the lower bound is applied correctly, we report the error rate of the neural \nand~in a manner that counts zero output forced by synaptic failure as an error.

\begin{figure}[htbp]
        \centering
        \includegraphics[width=0.95\linewidth]{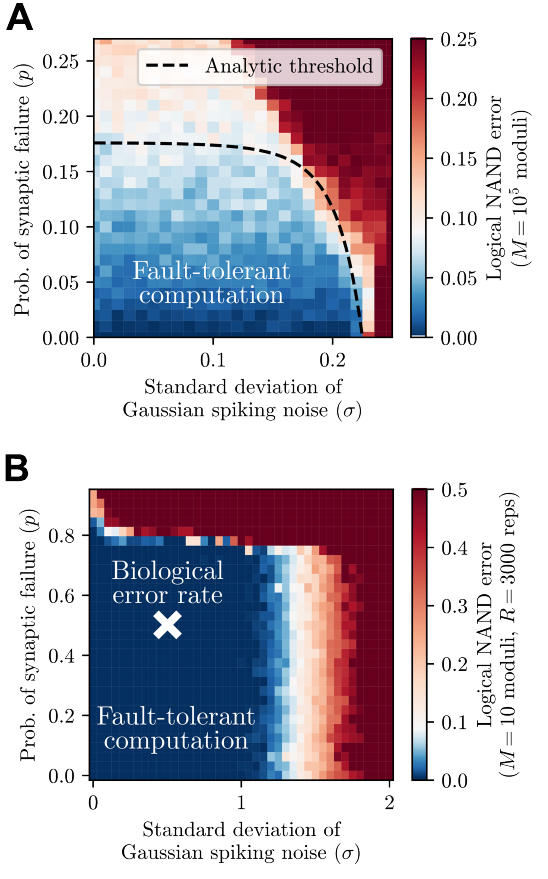}
        \caption{
        (\textbf{A}) The logical error of a neural \nand~gate using only the grid code. 
        Also plotted is an analytical fault-tolerance threshold corresponding to \nand~error probability \(\epsilon_0 \approx 0.09\) required to achieve an arbitrarily low logical error using the optimal \nand~fault-tolerance construction of Ref.~\cite{evans1998on-the-maximum}.
        The region in blue supports fault-tolerant computation, while the region in red suffers faulty computation.
        (\textbf{B}) Logical error of a neural implementation of the \nand~gate using the grid code (\(M=10\) moduli) concatenated with a repetition code (\(R=3000\) repetitions).
        Notably, the fault-tolerant regime where the error falls below \(\epsilon_0 \approx 0.09\) (blue) encompasses biological error rates (white cross).
    }
        \label{fig:threshold}
    \end{figure}

    We use both this bound and the expression for \(\epsilon(\sigma, p)\) (see \cref{sec:synaptic-failure-details}) to analytically determine a fault-tolerance threshold for \(p\) and \(\sigma\) at \(M = 10^5\) moduli. 
    We analytically plot the neural \nand~failure probability \(\epsilon(\sigma, p)\) in \cref{fig:threshold}a, as well as a contour (the dashed line) corresponding to the logical error being equal to the aforementioned threshold \(\epsilon_0 = (3-\sqrt 7)/4\). 
    This plot indicates a region of \(\sigma, p\) with logical error \(\epsilon(\sigma, p) < \epsilon_0\), within which fault-tolerant computation is achievable, and a sharp transition to a region with \(\epsilon(\sigma, p) > \epsilon_0\) in which this fault-tolerant construction does not hold. 
    In aggregate then, by appealing to the universality of the \nand~gate, we have that for sufficiently small \(\sigma < \sigma_0\) and \(p < p_0\) (where \(\sigma_0\) and \(p_0\) may be determined by the contour of \cref{fig:threshold}a), our fault-tolerant neuron may achieve fault-tolerant computation with polylogarithmic overhead, thus achieving neural network fault-tolerance as introduced in \cref{sec:intro}.

    \subsection{Concatenating grid code on top of repetition code}
    \label{sec:repetition-concat}    
    Above, we constructed a neural \nand~that uses \(M\) moduli, where each modulus is stored without redundancy. 
    While our constructions above have assumed \(M=10^5\), the grid cells in the mammalian cortex contain far fewer moduli, i.e., \(M \sim 10\) \cite{fiete2008what}.
    However, in the biological setting, each modulus is itself encoded redundantly, with \(R \sim 10^3\text{ to }10^4\) repetitions of each modulus \cite{fiete2008what}. 
    This effectively concatenates the grid code on top of a repetition code, which provides another means by which to decrease the strength of the additive Gaussian noise.
    Roughly speaking, the central limit theorem reduces the variance \(\sigma^2\) to \(\sigma^2/R\), which can drastically reduce the number of moduli required to suppress noise (\cref{fig:booleans}c).
    To move towards a more biologically feasible setting, we examine a concatenation of the grid code on top of a repetition code.

    Considering the logical neuron in \cref{fig:logical_neuron_relu}a, the main modification is to replace each phase with \(R\) copies of the phase.
    Each successive layer then averages over the repetitions of the previous layer, correcting for the synaptic failure probability.
    For example, consider the neural \nand~gate with \(R\) copies of the first phase \(\{\theta_{1,i}^{(1)}\}_{i=1}^R\) and \(R\) copies of the second phase \(\{\theta_{1,i}^{(2)}\}_{i=1}^R\).
    In the absence of the repetition code, the phase \(\phi_1\) would be computed as \(\phi_1 = \theta_{1}^{(1)} + \theta_{2}^{(1)}\) as the neural \nand~gate uses logical weights equal to \(1\).
    With the repetition code and a synaptic failure probability \(p\), we instead choose 
    \begin{align}
        \phi_{1,j} = \frac{1}{R}\times\frac{1}{1-p}\sum_{i=1}^R \theta_{1,i}^{(1)} + \theta_{1,i}^{(2)}
    \end{align}
    which may be implemented by selecting weights \(\frac{1}{R}\times\frac{1}{1-p}\), where the factor \(\frac{1}{1-p}\) accommodates for synaptic failure. 
    This ensures that i.i.d. sampling of the Gaussian noise over \(R\) repetitions will reduce the variance from \(\sigma^2\) to \(\sigma^2/R\).

    The only remaining modification is to use a step function in the encoder to perform a majority vote over repetitions in the final layer.
    The goal is to ensure that the character of the noise remains the same after decoding, i.e., the noise after error correction should be describable as a combination of logical bit flips and continuous Gaussian noise. 
    As in \cref{sec:gaussian-reliability}, consider a set of three codewords \(x_1 = 0\), \(x_2 = a\), and \(x_3 = 2a\), interpreting \(x_1\) as False and \(x_2\) as True.
    In the previously studied construction where \(R = 1\), the outputs of the error corrected \nand~gate corresponding to \(\phi_j\) are simply multiplied by weights \(1/\lambda_j\), \(1/\lambda_j\), and \(0\) respectively (see \cref{eq:boolean_codeword_1,eq:boolean_codeword_2} and \cref{fig:booleans}a). 
    In addition to rescaling by \(\frac{1}{1-p}\) to account for synaptic failures, we include an extra discretization step in the encoding stage of error correction (as in \cref{fig:logical_neuron_relu}a).
    This is accomplished by choosing \(e(x_k)\) to be a step function in the encoder of \cref{eq:nenc}.
    If \(x_1\) is recovered by the decoder, we re-encode \(e(x_1) = 0\);
    and if \(x_2\) is recovered by the decoder, we re-encode \(e(x_2) = a / (1 - p)\).
    Since the weight associated with \(x_3\) is zero, the decoding neuron corresponding to \(x_3\) is not connected to the following layer of neurons.

    We conduct numerical experiments on this neural \nand~gate with redundantly encoded moduli, using \(M=10\) moduli and \(R=3\times 10^3\) repetitions to remain in the biologically relevant regime. 
    As before, errors are biased due to synaptic failure setting neurons to zero; hence, the threshold of Evans and Pippenger places a lower bound on the true threshold of the neural \nand, where zero output incurred by synaptic failure is appropriately counted as a logical error.
    Since the central limit theorem performs poorly on the small number of moduli \(M=10\) here, an analytic expression like that for \(\epsilon(\sigma, p) \) (\cref{eq:psyn}) is a poor approximation. 
    Instead, we numerically estimate the threshold as the contour where the logical \nand~error crosses the Evans and Pippenger threshold \(\epsilon_0 = (3 - \sqrt{7}) / 4 \approx 0.09\). 
    We show results in \cref{fig:threshold}b, with the threshold contour depicted as the white boundary separating the blue region (which represents fault-tolerant computation) and the red region (which represents faulty computation). 
    This indicates approximate thresholds \(\sigma_0 \approx 1.4\) and \(p_0 \approx 0.7\).
    Notably, the fault-tolerant regime encompasses the observed biological error rates (depicted as a white cross) of \(\sigma \approx 0.5\) (given a mean of approximately 0.5, due to random outputs in \([0, 1]\)) and \(p \approx 0.5\)~\cite{stevens1994changes,hessler1993the-probability,softky1993the-highly}, thus suggesting that the grid code augmented with a repetition code suffices to enable reliable computation in faulty organisms.

  \section{Concluding remarks}
  \label{sec:conclusion}

    In this work, we have demonstrated fault-tolerant constructions for neural networks subject to synaptic failure (\cref{sec:digital-ft}) and additive Gaussian noise (\cref{sec:logical-neuron}). 
    While synaptic failure is a digital error and may be treated with a traditional repetition code, Gaussian noise represents an analog error, which we treat using the more sophisticated grid code that emerged from studies of the mammalian cortex. 
    We have further used these constructions to build neural networks that can reliably implement any Boolean formula in the presence of both errors modes (\cref{sec:demonstration-of-computation}). 
    In particular, for sufficiently small synaptic failure probability \(p < p_0\) and Gaussian noise standard deviation \(\sigma < \sigma_0\), our construction enables the computation of arbitrary Boolean formulas (and thus arbitrary computation) with only polylogarithmic overhead, thus achieving neural network fault-tolerance as introduced in \cref{sec:intro}. 
    These results ultimately describe a phase transition from faulty neural computation into fault-tolerant neural computation.
    
    Our analyses only place a lower bound on the fault-tolerance threshold of neural computation; a more effective neural fault-tolerant construction may be exist.
    In particular, while the neural network fault-tolerance theorem is phrased in terms of digital Boolean gates composed of analog neurons, the fault-tolerant neural network size requirement of \cref{sec:ft-gaussian-nn} (\cref{eq:m}) holds for a general construction of neural networks with Gaussian-distributed weights.
    This standard form of artificial neural networks provides a more direct analog approach to computation without introducing logical digital gates, and it may ultimately realize a more efficient path towards a threshold for the fault-tolerant phase of neural computation.

    Framed against the slowing pace of Moore's Law and increasingly prohibitive energy costs of deep learning~\cite{brown2020language,jumper2021highly}, the remarkable efficiency of biological computation places central importance on a deep understanding of noisy analog systems.
    The brain is a canonical example of a noisy analog system that is more energy-efficient than traditional faultless computation.
    By demonstrating the existence of fault-tolerant neural networks, our work provides a concrete path towards leveraging the favorable properties of such analog neural networks in a neuromorphic setting~\cite{indiveri2011neuromorphic,esser2016convolutional,wang2018fully}.
    These results may also find use in novel hardware for machine learning acceleration, such as optical computing~\cite{mcmahon2023physics} and thermodynamic computing~\cite{conte2019thermodynamic}, which may achieve more resource-efficient computations at the expense of increased error. 
    Our findings are suggestive of the power of naturally occurring error-correcting mechanisms: while the presence of fault-tolerant computation in the brain remains uncertain without experimental verification, we conclude that observed neural error correction codes are theoretically capable of achieving arbitrarily reliable computation.

\begin{acknowledgements}
    AZ acknowledges support from the Hertz Foundation, and the Department of Defense through the National Defense Science and Engineering Graduate Fellowship Program.
    AKT acknowledges support from the Natural Sciences and Engineering Research Council of Canada (NSERC) [PGSD3-545841-2020].
    MT acknowledges support from the Rothberg Family Fund for Cognitive Science.
    ILC, AKT, and MT acknowledge support in part from the Institute for Artificial Intelligence and Fundamental Interactions (IAIFI) through NSF Grant No. PHY-2019786.
\end{acknowledgements}

\bibliography{references}

\begin{thebibliography}{34}%
\makeatletter
\providecommand \@ifxundefined [1]{%
 \@ifx{#1\undefined}
}%
\providecommand \@ifnum [1]{%
 \ifnum #1\expandafter \@firstoftwo
 \else \expandafter \@secondoftwo
 \fi
}%
\providecommand \@ifx [1]{%
 \ifx #1\expandafter \@firstoftwo
 \else \expandafter \@secondoftwo
 \fi
}%
\providecommand \natexlab [1]{#1}%
\providecommand \enquote  [1]{``#1''}%
\providecommand \bibnamefont  [1]{#1}%
\providecommand \bibfnamefont [1]{#1}%
\providecommand \citenamefont [1]{#1}%
\providecommand \href@noop [0]{\@secondoftwo}%
\providecommand \href [0]{\begingroup \@sanitize@url \@href}%
\providecommand \@href[1]{\@@startlink{#1}\@@href}%
\providecommand \@@href[1]{\endgroup#1\@@endlink}%
\providecommand \@sanitize@url [0]{\catcode `\\12\catcode `\$12\catcode
  `\&12\catcode `\#12\catcode `\^12\catcode `\_12\catcode `\%12\relax}%
\providecommand \@@startlink[1]{}%
\providecommand \@@endlink[0]{}%
\providecommand \url  [0]{\begingroup\@sanitize@url \@url }%
\providecommand \@url [1]{\endgroup\@href {#1}{\urlprefix }}%
\providecommand \urlprefix  [0]{URL }%
\providecommand \Eprint [0]{\href }%
\providecommand \doibase [0]{http://dx.doi.org/}%
\providecommand \selectlanguage [0]{\@gobble}%
\providecommand \bibinfo  [0]{\@secondoftwo}%
\providecommand \bibfield  [0]{\@secondoftwo}%
\providecommand \translation [1]{[#1]}%
\providecommand \BibitemOpen [0]{}%
\providecommand \bibitemStop [0]{}%
\providecommand \bibitemNoStop [0]{.\EOS\space}%
\providecommand \EOS [0]{\spacefactor3000\relax}%
\providecommand \BibitemShut  [1]{\csname bibitem#1\endcsname}%
\let\auto@bib@innerbib\@empty
\bibitem [{\citenamefont {von Neumann}(2016)}]{von-neumann2016probabilistic}%
  \BibitemOpen
  \bibfield  {author} {\bibinfo {author} {\bibfnamefont {J.}~\bibnamefont {von
  Neumann}},\ }\enquote {\bibinfo {title} {Probabilistic logics and the
  synthesis of reliable organisms from unreliable components},}\ in\ \href
  {\doibase doi:10.1515/9781400882618-003} {\emph {\bibinfo {booktitle}
  {Automata Studies. (AM-34), Volume 34}}},\ \bibinfo {editor} {edited by\
  \bibinfo {editor} {\bibfnamefont {C.~E.}\ \bibnamefont {Shannon}}\ and\
  \bibinfo {editor} {\bibfnamefont {J.}~\bibnamefont {McCarthy}}}\ (\bibinfo
  {publisher} {Princeton University Press},\ \bibinfo {year} {2016})\ pp.\
  \bibinfo {pages} {43--98}\BibitemShut {NoStop}%
\bibitem [{\citenamefont {Pippenger}(1985)}]{pippenger1985on-networks}%
  \BibitemOpen
  \bibfield  {author} {\bibinfo {author} {\bibfnamefont {N.}~\bibnamefont
  {Pippenger}},\ }\href {\doibase 10.1109/SFCS.1985.41} {\bibfield  {journal}
  {\bibinfo  {journal} {26th Annual Symposium on Foundations of Computer
  Science (SFCS 1985)}\ ,\ \bibinfo {pages} {30}} (\bibinfo {year}
  {1985})}\BibitemShut {NoStop}%
\bibitem [{\citenamefont {Hajek}\ and\ \citenamefont
  {Weller}(1991)}]{hajek1991on-the-maximum}%
  \BibitemOpen
  \bibfield  {author} {\bibinfo {author} {\bibfnamefont {B.}~\bibnamefont
  {Hajek}}\ and\ \bibinfo {author} {\bibfnamefont {T.}~\bibnamefont {Weller}},\
  }\href {\doibase 10.1109/18.75261} {\bibfield  {journal} {\bibinfo  {journal}
  {IEEE Transactions on Information Theory}\ }\textbf {\bibinfo {volume}
  {37}},\ \bibinfo {pages} {388} (\bibinfo {year} {1991})}\BibitemShut
  {NoStop}%
\bibitem [{\citenamefont {Evans}\ and\ \citenamefont
  {Pippenger}(1998)}]{evans1998on-the-maximum}%
  \BibitemOpen
  \bibfield  {author} {\bibinfo {author} {\bibfnamefont {W.}~\bibnamefont
  {Evans}}\ and\ \bibinfo {author} {\bibfnamefont {N.}~\bibnamefont
  {Pippenger}},\ }\href {\doibase 10.1109/18.669417} {\bibfield  {journal}
  {\bibinfo  {journal} {IEEE Transactions on Information Theory}\ }\textbf
  {\bibinfo {volume} {44}} (\bibinfo {year} {1998}),\
  10.1109/18.669417}\BibitemShut {NoStop}%
\bibitem [{\citenamefont {Evans}\ and\ \citenamefont
  {Schulman}(1999)}]{evans1999signal}%
  \BibitemOpen
  \bibfield  {author} {\bibinfo {author} {\bibfnamefont {W.}~\bibnamefont
  {Evans}}\ and\ \bibinfo {author} {\bibfnamefont {L.}~\bibnamefont
  {Schulman}},\ }\href {\doibase 10.1109/18.796377} {\bibfield  {journal}
  {\bibinfo  {journal} {IEEE Transactions on Information Theory}\ }\textbf
  {\bibinfo {volume} {45}},\ \bibinfo {pages} {2367} (\bibinfo {year}
  {1999})}\BibitemShut {NoStop}%
\bibitem [{\citenamefont {Gao}\ \emph {et~al.}(2005)\citenamefont {Gao},
  \citenamefont {Qi},\ and\ \citenamefont {Fortes}}]{gao2005bifurcations}%
  \BibitemOpen
  \bibfield  {author} {\bibinfo {author} {\bibfnamefont {J.}~\bibnamefont
  {Gao}}, \bibinfo {author} {\bibfnamefont {Y.}~\bibnamefont {Qi}}, \ and\
  \bibinfo {author} {\bibfnamefont {J.}~\bibnamefont {Fortes}},\ }\href
  {\doibase 10.1109/TNANO.2005.851289} {\bibfield  {journal} {\bibinfo
  {journal} {IEEE Transactions on Nanotechnology}\ }\textbf {\bibinfo {volume}
  {4}},\ \bibinfo {pages} {395} (\bibinfo {year} {2005})}\BibitemShut {NoStop}%
\bibitem [{\citenamefont {Shor}(1996)}]{shor1996fault-tolerant}%
  \BibitemOpen
  \bibfield  {author} {\bibinfo {author} {\bibfnamefont {P.}~\bibnamefont
  {Shor}},\ }in\ \href {\doibase 10.1109/SFCS.1996.548464} {\emph {\bibinfo
  {booktitle} {Proceedings of 37th Conference on Foundations of Computer
  Science}}}\ (\bibinfo {year} {1996})\ pp.\ \bibinfo {pages}
  {56--65}\BibitemShut {NoStop}%
\bibitem [{\citenamefont {Torres-Huitzil}\ and\ \citenamefont
  {Girau}(2017)}]{torres-huitzil2017fault}%
  \BibitemOpen
  \bibfield  {author} {\bibinfo {author} {\bibfnamefont {C.}~\bibnamefont
  {Torres-Huitzil}}\ and\ \bibinfo {author} {\bibfnamefont {B.}~\bibnamefont
  {Girau}},\ }\href {\doibase 10.1109/ACCESS.2017.2742698} {\bibfield
  {journal} {\bibinfo  {journal} {IEEE Access}\ }\textbf {\bibinfo {volume}
  {5}},\ \bibinfo {pages} {17322} (\bibinfo {year} {2017})}\BibitemShut
  {NoStop}%
\bibitem [{\citenamefont {Hafting}\ \emph {et~al.}(2005)\citenamefont
  {Hafting}, \citenamefont {Fyhn}, \citenamefont {Molden}, \citenamefont
  {Moser},\ and\ \citenamefont {Moser}}]{hafting2005microstructure}%
  \BibitemOpen
  \bibfield  {author} {\bibinfo {author} {\bibfnamefont {T.}~\bibnamefont
  {Hafting}}, \bibinfo {author} {\bibfnamefont {M.}~\bibnamefont {Fyhn}},
  \bibinfo {author} {\bibfnamefont {S.}~\bibnamefont {Molden}}, \bibinfo
  {author} {\bibfnamefont {M.-B.}\ \bibnamefont {Moser}}, \ and\ \bibinfo
  {author} {\bibfnamefont {E.~I.}\ \bibnamefont {Moser}},\ }\href
  {https://doi.org/10.1038/nature03721} {\bibfield  {journal} {\bibinfo
  {journal} {Nature}\ }\textbf {\bibinfo {volume} {436}},\ \bibinfo {pages}
  {801} (\bibinfo {year} {2005})}\BibitemShut {NoStop}%
\bibitem [{\citenamefont {Fiete}\ \emph {et~al.}(2008)\citenamefont {Fiete},
  \citenamefont {Burak},\ and\ \citenamefont {Brookings}}]{fiete2008what}%
  \BibitemOpen
  \bibfield  {author} {\bibinfo {author} {\bibfnamefont {I.~R.}\ \bibnamefont
  {Fiete}}, \bibinfo {author} {\bibfnamefont {Y.}~\bibnamefont {Burak}}, \ and\
  \bibinfo {author} {\bibfnamefont {T.}~\bibnamefont {Brookings}},\ }\href
  {https://doi.org/10.1523/JNEUROSCI.5684-07.2008} {\bibfield  {journal}
  {\bibinfo  {journal} {Journal of Neuroscience}\ }\textbf {\bibinfo {volume}
  {28}},\ \bibinfo {pages} {6858} (\bibinfo {year} {2008})}\BibitemShut
  {NoStop}%
\bibitem [{\citenamefont {Sreenivasan}\ and\ \citenamefont
  {Fiete}(2011)}]{sreenivasan2011grid}%
  \BibitemOpen
  \bibfield  {author} {\bibinfo {author} {\bibfnamefont {S.}~\bibnamefont
  {Sreenivasan}}\ and\ \bibinfo {author} {\bibfnamefont {I.}~\bibnamefont
  {Fiete}},\ }\href {\doibase 10.1038/nn.2901} {\bibfield  {journal} {\bibinfo
  {journal} {Nature Neuroscience}\ }\textbf {\bibinfo {volume} {14}},\ \bibinfo
  {pages} {1330} (\bibinfo {year} {2011})}\BibitemShut {NoStop}%
\bibitem [{\citenamefont {Stevens}\ and\ \citenamefont
  {Wang}(1994)}]{stevens1994changes}%
  \BibitemOpen
  \bibfield  {author} {\bibinfo {author} {\bibfnamefont {C.~F.}\ \bibnamefont
  {Stevens}}\ and\ \bibinfo {author} {\bibfnamefont {Y.}~\bibnamefont {Wang}},\
  }\href {\doibase 10.1038/371704a0} {\bibfield  {journal} {\bibinfo  {journal}
  {Nature}\ }\textbf {\bibinfo {volume} {371}},\ \bibinfo {pages} {704}
  (\bibinfo {year} {1994})}\BibitemShut {NoStop}%
\bibitem [{\citenamefont {Hessler}\ \emph {et~al.}(1993)\citenamefont
  {Hessler}, \citenamefont {Shirke},\ and\ \citenamefont
  {Malinow}}]{hessler1993the-probability}%
  \BibitemOpen
  \bibfield  {author} {\bibinfo {author} {\bibfnamefont {N.~A.}\ \bibnamefont
  {Hessler}}, \bibinfo {author} {\bibfnamefont {A.~M.}\ \bibnamefont {Shirke}},
  \ and\ \bibinfo {author} {\bibfnamefont {R.}~\bibnamefont {Malinow}},\ }\href
  {\doibase 10.1038/366569a0} {\bibfield  {journal} {\bibinfo  {journal}
  {Nature}\ }\textbf {\bibinfo {volume} {366}},\ \bibinfo {pages} {569}
  (\bibinfo {year} {1993})}\BibitemShut {NoStop}%
\bibitem [{\citenamefont {Softky}\ and\ \citenamefont
  {Koch}(1993)}]{softky1993the-highly}%
  \BibitemOpen
  \bibfield  {author} {\bibinfo {author} {\bibfnamefont {W.}~\bibnamefont
  {Softky}}\ and\ \bibinfo {author} {\bibfnamefont {C.}~\bibnamefont {Koch}},\
  }\href {\doibase 10.1523/JNEUROSCI.13-01-00334.1993} {\bibfield  {journal}
  {\bibinfo  {journal} {Journal of Neuroscience}\ }\textbf {\bibinfo {volume}
  {13}},\ \bibinfo {pages} {334} (\bibinfo {year} {1993})}\BibitemShut
  {NoStop}%
\bibitem [{\citenamefont {McCulloch}\ and\ \citenamefont
  {Pitts}(1943)}]{mcculloch1943a-logical}%
  \BibitemOpen
  \bibfield  {author} {\bibinfo {author} {\bibfnamefont {W.~S.}\ \bibnamefont
  {McCulloch}}\ and\ \bibinfo {author} {\bibfnamefont {W.}~\bibnamefont
  {Pitts}},\ }\href {\doibase 10.1007/BF02478259} {\bibfield  {journal}
  {\bibinfo  {journal} {The Bulletin of Mathematical Biophysics}\ }\textbf
  {\bibinfo {volume} {5}},\ \bibinfo {pages} {115} (\bibinfo {year}
  {1943})}\BibitemShut {NoStop}%
\bibitem [{\citenamefont {Hornik}\ \emph {et~al.}(1989)\citenamefont {Hornik},
  \citenamefont {Stinchcombe},\ and\ \citenamefont
  {White}}]{hornik1989multilayer}%
  \BibitemOpen
  \bibfield  {author} {\bibinfo {author} {\bibfnamefont {K.}~\bibnamefont
  {Hornik}}, \bibinfo {author} {\bibfnamefont {M.}~\bibnamefont {Stinchcombe}},
  \ and\ \bibinfo {author} {\bibfnamefont {H.}~\bibnamefont {White}},\
  }\href@noop {} {\bibfield  {journal} {\bibinfo  {journal} {Neural Networks}\
  }\textbf {\bibinfo {volume} {2}},\ \bibinfo {pages} {359} (\bibinfo {year}
  {1989})}\BibitemShut {NoStop}%
\bibitem [{\citenamefont {LeCun}\ \emph {et~al.}(2015)\citenamefont {LeCun},
  \citenamefont {Bengio},\ and\ \citenamefont {Hinton}}]{lecun2015deep}%
  \BibitemOpen
  \bibfield  {author} {\bibinfo {author} {\bibfnamefont {Y.}~\bibnamefont
  {LeCun}}, \bibinfo {author} {\bibfnamefont {Y.}~\bibnamefont {Bengio}}, \
  and\ \bibinfo {author} {\bibfnamefont {G.}~\bibnamefont {Hinton}},\ }\href
  {https://doi.org/10.1038/nature14539} {\bibfield  {journal} {\bibinfo
  {journal} {Nature}\ }\textbf {\bibinfo {volume} {521}},\ \bibinfo {pages}
  {436} (\bibinfo {year} {2015})}\BibitemShut {NoStop}%
\bibitem [{\citenamefont {Neti}\ \emph {et~al.}(1992)\citenamefont {Neti},
  \citenamefont {Schneider},\ and\ \citenamefont {Young}}]{neti1992maximally}%
  \BibitemOpen
  \bibfield  {author} {\bibinfo {author} {\bibfnamefont {C.}~\bibnamefont
  {Neti}}, \bibinfo {author} {\bibfnamefont {M.}~\bibnamefont {Schneider}}, \
  and\ \bibinfo {author} {\bibfnamefont {E.}~\bibnamefont {Young}},\ }\href
  {\doibase 10.1109/72.105414} {\bibfield  {journal} {\bibinfo  {journal} {IEEE
  Transactions on Neural Networks}\ }\textbf {\bibinfo {volume} {3}},\ \bibinfo
  {pages} {14} (\bibinfo {year} {1992})}\BibitemShut {NoStop}%
\bibitem [{\citenamefont {El~Mhamdi}\ and\ \citenamefont
  {Guerraoui}(2017)}]{mhamdi2017when}%
  \BibitemOpen
  \bibfield  {author} {\bibinfo {author} {\bibfnamefont {E.~M.}\ \bibnamefont
  {El~Mhamdi}}\ and\ \bibinfo {author} {\bibfnamefont {R.}~\bibnamefont
  {Guerraoui}},\ }in\ \href@noop {} {\emph {\bibinfo {booktitle} {2017 IEEE
  International Parallel and Distributed Processing Symposium (IPDPS)}}}\
  (\bibinfo {year} {2017})\ pp.\ \bibinfo {pages} {1028--1037}\BibitemShut
  {NoStop}%
\bibitem [{\citenamefont {Liu}\ \emph {et~al.}(2019)\citenamefont {Liu},
  \citenamefont {Wen}, \citenamefont {Jiang}, \citenamefont {Wang},
  \citenamefont {Yang},\ and\ \citenamefont {Quan}}]{liu2019a-fault-tolerant}%
  \BibitemOpen
  \bibfield  {author} {\bibinfo {author} {\bibfnamefont {T.}~\bibnamefont
  {Liu}}, \bibinfo {author} {\bibfnamefont {W.}~\bibnamefont {Wen}}, \bibinfo
  {author} {\bibfnamefont {L.}~\bibnamefont {Jiang}}, \bibinfo {author}
  {\bibfnamefont {Y.}~\bibnamefont {Wang}}, \bibinfo {author} {\bibfnamefont
  {C.}~\bibnamefont {Yang}}, \ and\ \bibinfo {author} {\bibfnamefont
  {G.}~\bibnamefont {Quan}},\ }in\ \href@noop {} {\emph {\bibinfo {booktitle}
  {2019 56th ACM/IEEE Design Automation Conference (DAC)}}}\ (\bibinfo {year}
  {2019})\ pp.\ \bibinfo {pages} {1--6}\BibitemShut {NoStop}%
\bibitem [{\citenamefont {Sequin}\ and\ \citenamefont
  {Clay}(1990)}]{sequin1990fault}%
  \BibitemOpen
  \bibfield  {author} {\bibinfo {author} {\bibfnamefont {C.}~\bibnamefont
  {Sequin}}\ and\ \bibinfo {author} {\bibfnamefont {R.}~\bibnamefont {Clay}},\
  }in\ \href {\doibase 10.1109/IJCNN.1990.137651} {\emph {\bibinfo {booktitle}
  {1990 IJCNN International Joint Conference on Neural Networks}}}\ (\bibinfo
  {year} {1990})\ pp.\ \bibinfo {pages} {703--708 vol.1}\BibitemShut {NoStop}%
\bibitem [{\citenamefont {Ruckert}\ \emph {et~al.}(1989)\citenamefont
  {Ruckert}, \citenamefont {Kreuzer}, \citenamefont {Tryba},\ and\
  \citenamefont {Goser}}]{ruckert1989fault-tolerance}%
  \BibitemOpen
  \bibfield  {author} {\bibinfo {author} {\bibfnamefont {U.}~\bibnamefont
  {Ruckert}}, \bibinfo {author} {\bibfnamefont {I.}~\bibnamefont {Kreuzer}},
  \bibinfo {author} {\bibfnamefont {V.}~\bibnamefont {Tryba}}, \ and\ \bibinfo
  {author} {\bibfnamefont {K.}~\bibnamefont {Goser}},\ }in\ \href {\doibase
  10.1109/CMPEUR.1989.93343} {\emph {\bibinfo {booktitle} {Proceedings. {VLSI}
  and Computer Peripherals. {COMPEURO} 89}}}\ (\bibinfo {year} {1989})\ pp.\
  \bibinfo {pages} {1/52--1/55}\BibitemShut {NoStop}%
\bibitem [{\citenamefont {von Neumann}(1956)}]{von-neumann1956probabilistic}%
  \BibitemOpen
  \bibfield  {author} {\bibinfo {author} {\bibfnamefont {J.}~\bibnamefont {von
  Neumann}},\ }\href@noop {} {\bibfield  {journal} {\bibinfo  {journal}
  {Automata Studies}\ }\textbf {\bibinfo {volume} {34}},\ \bibinfo {pages} {43}
  (\bibinfo {year} {1956})}\BibitemShut {NoStop}%
\bibitem [{\citenamefont {Evans}\ and\ \citenamefont
  {Schulman}(2003)}]{evans2003on-the-maximum}%
  \BibitemOpen
  \bibfield  {author} {\bibinfo {author} {\bibfnamefont {W.}~\bibnamefont
  {Evans}}\ and\ \bibinfo {author} {\bibfnamefont {L.}~\bibnamefont
  {Schulman}},\ }\href {\doibase 10.1109/TIT.2003.818405} {\bibfield  {journal}
  {\bibinfo  {journal} {IEEE Transactions on Information Theory}\ }\textbf
  {\bibinfo {volume} {49}},\ \bibinfo {pages} {3094} (\bibinfo {year}
  {2003})}\BibitemShut {NoStop}%
\bibitem [{\citenamefont {Winograd}\ and\ \citenamefont
  {Cowan}(1963)}]{winograd1963reliable}%
  \BibitemOpen
  \bibfield  {author} {\bibinfo {author} {\bibfnamefont {S.}~\bibnamefont
  {Winograd}}\ and\ \bibinfo {author} {\bibfnamefont {J.~D.}\ \bibnamefont
  {Cowan}},\ }\href@noop {} {\emph {\bibinfo {title} {Reliable Computation in
  the Presence of Noise}}}\ (\bibinfo  {publisher} {MIT Press Cambridge,
  Mass.},\ \bibinfo {year} {1963})\BibitemShut {NoStop}%
\bibitem [{\citenamefont {Nielsen}\ and\ \citenamefont
  {Chuang}(2010)}]{nielsen2010quantum}%
  \BibitemOpen
  \bibfield  {author} {\bibinfo {author} {\bibfnamefont {M.~A.}\ \bibnamefont
  {Nielsen}}\ and\ \bibinfo {author} {\bibfnamefont {I.~L.}\ \bibnamefont
  {Chuang}},\ }\href@noop {} {\emph {\bibinfo {title} {Quantum computation and
  quantum information}}}\ (\bibinfo  {publisher} {Cambridge university press},\
  \bibinfo {year} {2010})\BibitemShut {NoStop}%
\bibitem [{\citenamefont {Goblick}(1965)}]{goblick1965theoretical}%
  \BibitemOpen
  \bibfield  {author} {\bibinfo {author} {\bibfnamefont {T.}~\bibnamefont
  {Goblick}},\ }\href {\doibase 10.1109/TIT.1965.1053821} {\bibfield  {journal}
  {\bibinfo  {journal} {IEEE Transactions on Information Theory}\ }\textbf
  {\bibinfo {volume} {11}},\ \bibinfo {pages} {558} (\bibinfo {year}
  {1965})}\BibitemShut {NoStop}%
\bibitem [{\citenamefont {Brown}\ \emph {et~al.}(2020)\citenamefont {Brown},
  \citenamefont {Mann}, \citenamefont {Ryder}, \citenamefont {Subbiah},
  \citenamefont {Kaplan} \emph {et~al.}}]{brown2020language}%
  \BibitemOpen
  \bibfield  {author} {\bibinfo {author} {\bibfnamefont {T.}~\bibnamefont
  {Brown}}, \bibinfo {author} {\bibfnamefont {B.}~\bibnamefont {Mann}},
  \bibinfo {author} {\bibfnamefont {N.}~\bibnamefont {Ryder}}, \bibinfo
  {author} {\bibfnamefont {M.}~\bibnamefont {Subbiah}}, \bibinfo {author}
  {\bibfnamefont {J.~D.}\ \bibnamefont {Kaplan}},  \emph {et~al.},\ }in\
  \href@noop {} {\emph {\bibinfo {booktitle} {Advances in Neural Information
  Processing Systems}}},\ Vol.~\bibinfo {volume} {33},\ \bibinfo {editor}
  {edited by\ \bibinfo {editor} {\bibfnamefont {H.}~\bibnamefont {Larochelle}},
  \bibinfo {editor} {\bibfnamefont {M.}~\bibnamefont {Ranzato}}, \bibinfo
  {editor} {\bibfnamefont {R.}~\bibnamefont {Hadsell}}, \bibinfo {editor}
  {\bibfnamefont {M.}~\bibnamefont {Balcan}}, \ and\ \bibinfo {editor}
  {\bibfnamefont {H.}~\bibnamefont {Lin}}}\ (\bibinfo  {publisher} {Curran
  Associates, Inc.},\ \bibinfo {year} {2020})\ pp.\ \bibinfo {pages}
  {1877--1901}\BibitemShut {NoStop}%
\bibitem [{\citenamefont {Jumper}\ \emph {et~al.}(2021)\citenamefont {Jumper},
  \citenamefont {Evans}, \citenamefont {Pritzel} \emph
  {et~al.}}]{jumper2021highly}%
  \BibitemOpen
  \bibfield  {author} {\bibinfo {author} {\bibfnamefont {J.}~\bibnamefont
  {Jumper}}, \bibinfo {author} {\bibfnamefont {R.}~\bibnamefont {Evans}},
  \bibinfo {author} {\bibfnamefont {A.}~\bibnamefont {Pritzel}},  \emph
  {et~al.},\ }\href@noop {} {\bibfield  {journal} {\bibinfo  {journal}
  {Nature}\ }\textbf {\bibinfo {volume} {596}},\ \bibinfo {pages} {583}
  (\bibinfo {year} {2021})}\BibitemShut {NoStop}%
\bibitem [{\citenamefont {Indiveri}\ \emph {et~al.}(2011)\citenamefont
  {Indiveri}, \citenamefont {Linares-Barranco}, \citenamefont {Hamilton},
  \citenamefont {van Schaik}, \citenamefont {Etienne-Cummings} \emph
  {et~al.}}]{indiveri2011neuromorphic}%
  \BibitemOpen
  \bibfield  {author} {\bibinfo {author} {\bibfnamefont {G.}~\bibnamefont
  {Indiveri}}, \bibinfo {author} {\bibfnamefont {B.}~\bibnamefont
  {Linares-Barranco}}, \bibinfo {author} {\bibfnamefont {T.}~\bibnamefont
  {Hamilton}}, \bibinfo {author} {\bibfnamefont {A.}~\bibnamefont {van
  Schaik}}, \bibinfo {author} {\bibfnamefont {R.}~\bibnamefont
  {Etienne-Cummings}},  \emph {et~al.},\ }\href {\doibase
  10.3389/fnins.2011.00073} {\bibfield  {journal} {\bibinfo  {journal}
  {Frontiers in Neuroscience}\ }\textbf {\bibinfo {volume} {5}},\ \bibinfo
  {pages} {73} (\bibinfo {year} {2011})}\BibitemShut {NoStop}%
\bibitem [{\citenamefont {Esser}\ \emph {et~al.}(2016)\citenamefont {Esser},
  \citenamefont {Merolla}, \citenamefont {Arthur}, \citenamefont {Cassidy},
  \citenamefont {Appuswamy} \emph {et~al.}}]{esser2016convolutional}%
  \BibitemOpen
  \bibfield  {author} {\bibinfo {author} {\bibfnamefont {S.~K.}\ \bibnamefont
  {Esser}}, \bibinfo {author} {\bibfnamefont {P.~A.}\ \bibnamefont {Merolla}},
  \bibinfo {author} {\bibfnamefont {J.~V.}\ \bibnamefont {Arthur}}, \bibinfo
  {author} {\bibfnamefont {A.~S.}\ \bibnamefont {Cassidy}}, \bibinfo {author}
  {\bibfnamefont {R.}~\bibnamefont {Appuswamy}},  \emph {et~al.},\ }\href
  {\doibase 10.1073/pnas.1604850113} {\bibfield  {journal} {\bibinfo  {journal}
  {Proceedings of the National Academy of Sciences}\ }\textbf {\bibinfo
  {volume} {113}},\ \bibinfo {pages} {11441} (\bibinfo {year}
  {2016})}\BibitemShut {NoStop}%
\bibitem [{\citenamefont {Wang}\ \emph {et~al.}(2018)\citenamefont {Wang},
  \citenamefont {Joshi}, \citenamefont {Savel'ev}, \citenamefont {Song},
  \citenamefont {Midya} \emph {et~al.}}]{wang2018fully}%
  \BibitemOpen
  \bibfield  {author} {\bibinfo {author} {\bibfnamefont {Z.}~\bibnamefont
  {Wang}}, \bibinfo {author} {\bibfnamefont {S.}~\bibnamefont {Joshi}},
  \bibinfo {author} {\bibfnamefont {S.}~\bibnamefont {Savel'ev}}, \bibinfo
  {author} {\bibfnamefont {W.}~\bibnamefont {Song}}, \bibinfo {author}
  {\bibfnamefont {R.}~\bibnamefont {Midya}},  \emph {et~al.},\ }\href {\doibase
  10.1038/s41928-018-0023-2} {\bibfield  {journal} {\bibinfo  {journal} {Nature
  Electronics}\ }\textbf {\bibinfo {volume} {1}},\ \bibinfo {pages} {137}
  (\bibinfo {year} {2018})}\BibitemShut {NoStop}%
\bibitem [{\citenamefont {McMahon}(2023)}]{mcmahon2023physics}%
  \BibitemOpen
  \bibfield  {author} {\bibinfo {author} {\bibfnamefont {P.~L.}\ \bibnamefont
  {McMahon}},\ }\href {\doibase 10.1038/s42254-023-00645-5} {\bibfield
  {journal} {\bibinfo  {journal} {Nature Reviews Physics}\ }\textbf {\bibinfo
  {volume} {5}},\ \bibinfo {pages} {717–734} (\bibinfo {year}
  {2023})}\BibitemShut {NoStop}%
\bibitem [{\citenamefont {Conte}\ \emph {et~al.}(2019)\citenamefont {Conte},
  \citenamefont {DeBenedictis}, \citenamefont {Ganesh}, \citenamefont {Hylton},
  \citenamefont {Strachan} \emph {et~al.}}]{conte2019thermodynamic}%
  \BibitemOpen
  \bibfield  {author} {\bibinfo {author} {\bibfnamefont {T.}~\bibnamefont
  {Conte}}, \bibinfo {author} {\bibfnamefont {E.}~\bibnamefont {DeBenedictis}},
  \bibinfo {author} {\bibfnamefont {N.}~\bibnamefont {Ganesh}}, \bibinfo
  {author} {\bibfnamefont {T.}~\bibnamefont {Hylton}}, \bibinfo {author}
  {\bibfnamefont {J.~P.}\ \bibnamefont {Strachan}},  \emph {et~al.},\
  }\href@noop {} {\enquote {\bibinfo {title} {Thermodynamic computing},}\ }
  (\bibinfo {year} {2019}),\ \Eprint {http://arxiv.org/abs/1911.01968}
  {arXiv:1911.01968 [cs.CY]} \BibitemShut {NoStop}%
\end{thebibliography}%
\bibliographystyle{apsrev4-1}

\clearpage

\appendix

\section{Comparison of repetition for discrete versus analog fault-tolerance}
\label{sec:AnalogRepetitionScaling}
    While the repetition code is sufficient to arrive at digital fault-tolerance when subject to digital errors, such as bit flips or synaptic failure, it is insufficient for analog computation in the presence of additive Gaussian noise.
    Key to this is the \(\mathcal{O}(\operatorname{polylog}(1/\epsilon))\) scaling with respect to the desired output error rate \(\epsilon\) in the definition of fault-tolerance.
    For Boolean (and more generally discrete) random variables, suffering from i.i.d. bit-flip errors at a rate \(p < 1/2\), a repetition code of size \(M\) reduces errors exponentially as \(\sim p^{M}\).
    Given a target error rate \(\epsilon\), it is sufficient to choose
    \begin{equation}\label{eq:rep_cod_size}
        M \sim \frac{\log \frac{1}{\epsilon}}{\log \frac{1}{p}}.
    \end{equation}
    
    For a circuit of \(N\) gates, an overall error of \(\epsilon\) could be achieved by demanding individual gate errors \(\epsilon/N\) as per the union bound. Inserting this desired error rate into \cref{eq:rep_cod_size}, and using results of the concatenation scheme described in \cref{sec:ft-circuits}, we find that this translates to the desired \(\mathcal{O}(N \text{polylog}(N/\epsilon))\) scaling in the definition of fault-tolerance, so long as the error rate is below a threshold \(p_0\) that is dependent on the details of the error correcting circuit.

    For analog variables, the repetition code does not suppress errors strongly enough to achieve this scaling.
    For additive Gaussian noise with standard deviation \(\sigma\), a repetition code of size \(M\) suppresses errors not exponentially in \(M\), but only as \(\sim \sigma / \sqrt{M}\).
    For a target standard deviation \(\epsilon\), the code size is required to scale as
    \begin{equation}
        M \sim \left(\frac{\sigma}{\epsilon}\right)^2.
    \end{equation}
    Analog computation using the repetition code would require an asymptotic lower bound of \(\Omega(\operatorname{poly}(1/\epsilon))\) resources, and thus does not meet our definition of fault-tolerance.
    In order to achieve analog fault-tolerance, we must make use of a stronger error correction code, such as the grid code utillzed in this work.

\section{Detailed analysis of reliability in the presence of Gaussian noise and synaptic failure}
\label{sec:synaptic-failure-details}
    In this Appendix, we expand on the analysis in \cref{sec:gaussian-and-synaptic-reliability} for the fully general case that takes into account both Gaussian errors and synaptic failure.

    With the analysis for Gaussian failures worked out in \cref{sec:gaussian-reliability}, we proceed to consider the effect of synaptic failure for each possible type of synapse in the logical neuron of~\cref{fig:logical_neuron_relu}a.
    The goal is to find an upper bound on the probability that the logical \nand~fails, corresponding to a lower bound on the threshold for synaptic failure.

    First, considering the synapses from the decoder neurons \(x_i\) to the new logical phases \(\phi_j'\) (i.e., the final layer of \cref{fig:logical_neuron_relu}a), a failed synapse may originate from the correct decoder neuron or an incorrect decoder neuron.
    We ignore the failed synapse from an incorrect decoding, consistent with upper-bounding the failure probability.
    If the correct decoding fails, the encoded phase may not fire.
    In the application of logical weights to the logical phase of the next neuron (i.e., the first layer of \cref{fig:logical_neuron_relu}a), the synapse with a logical weight \(a_i\) may similarly fail.
    The two phenomena of a correct decoder synapse failing and a logical weight synapse failing produce the same outcome: an input phase \(\theta_i^{(1,2)}\) may fail.
    The effect of only a single input phase (e.g. \(\theta_i^{(1)}\)) failing is different from the effect of both input phases failing (i.e., \(\theta_i^{(1)}\) and \(\theta_i^{(2)}\)).
    If one input phase fails, the logical phase \(\phi_i\) assumes a uniformly random value from 0 to 1.
    This has no impact on \(f_\nand(k \neq k^*)\), but it reduces the mean of \(f_\nand(k = k^*)\) by removing one of the moduli and requires adjustment of the standard deviation by the inclusion of a random phase.
    If both input phases fail, the logical phase does not fire.
    Hence, one of the moduli is removed from both \(f_\nand(k \neq k^*)\) and \(f_\nand(k = k^*)\).
    In total, \(4Mp(1-p)\) single input phases are expected to fail and \(2Mp^2\) double input phases are expected to fail.

    Next, consider the synapses into and out of the \(\sin 2\pi\phi_i\) and \(\cos 2\pi\phi_i\) neurons.
    Here, we also find two cases: if there is a failure of a single sine or cosine, the original distribution must be compensated by the remaining sine or cosine of the phase; if there is a failure of both, the modulus is removed entirely.
    In expectation, \(2Mp(1-p)\) failures are expected for the former effect (for each of sine and cosine), and \(2Mp^2\) failures are expected for the latter.
    By adding each of the failure modes independently, we place an upper bound on logical failure due to double-counting failures that happen sequentially in the network.
    
    To obtain \(f_\nand'(k \neq k^*)\) and \(f_\nand'(k = k^*)\), we repeat the noisy logical neuron analysis of \cref{eq:fk0,eq:fnk0} for the neural \nand~ construction including possibility of synaptic failure detailed above.
    We make the same assumptions as for \cref{eq:fk0-nand,eq:fnk0-nand}, namely logical weights \(a_i = 1\) and \(S = 3\) decoder neurons as per the neural \nand~gate construction.
    Assuming a large number of moduli \(M \gg 1\) and applying the central limit theorem, we obtain
    \begin{equation}
    \label{eq:fk0p-nand}
        \begin{aligned}
        &f_\nand'(k = k^*) = \\
        &\quad \mathcal{N} \bigg(M \cdot \frac{e^{-6 \pi ^2 \sigma ^2} \operatorname{erf}\left(\sqrt{2} \pi  \sigma \right)^6}{2^9 \pi ^3 \sigma ^6} \cdot (2 (p-3) p+1), \\
        &\qquad \quad \qquad\sqrt{\sigma ^2-\frac{1}{2} M ((2 p (p+1)-1) \left(2 \sigma ^2+1\right) - \zeta')} \bigg),
        \end{aligned}
    \end{equation}
    where
    \begin{equation}
    \begin{aligned}
        \zeta' &= 2^{-18} \pi^{-6} \sigma^{-12} \Bigg[(p (3 p-7)+1) e^{-24 \pi ^2 \sigma^2} \times \\
        &\quad \left(e^{12 \pi^2 \sigma ^2} \operatorname{erf}\left(\sqrt{2} \pi  \sigma \right)^{12}-4 \pi ^3 \sigma ^6 \operatorname{erf}\left(2 \sqrt{2} \pi  \sigma \right)^6\right) \Bigg],
    \end{aligned}
    \end{equation}
    and
    \begin{equation}
    \label{eq:fnk0p-nand}
    \begin{aligned}
        f_\nand'(k \neq k^*) &= \mathcal{N}\bigg(0, \\
        &\quad \sqrt{\sigma ^2-\frac{1}{2} M (2 p (p+1)-1) \left(2 \sigma ^2+1\right)}\bigg).
    \end{aligned}
    \end{equation}

    Given \cref{eq:fk0p-nand,eq:fnk0p-nand}, we may evaluate the probability of successful decoding.
    As explained in \cref{sec:gaussian-and-synaptic-reliability}, we use a threshold non-linearity for decoding which is more biologically plausible than the alternative winner-take-all dynamics due to its locality.
    Choosing a threshold value of \(c\), a correct decoding then requires the correct neuron sampled from \(f_\nand'(k = k^*)\) (\cref{eq:fk0p-nand}) to exceed \(c\) and the two incorrect neurons sampled from \(f_\nand'(k \neq k^*)\) (\cref{eq:fnk0p-nand}) to lie below \(c\), i.e.
    \begin{equation}
        \begin{aligned}
        1 - &\epsilon(c; \sigma, p) := \\
        &\operatorname{Pr}[f_\nand'(k = k^*) > c] \times \operatorname{Pr}[f_\nand'(k \neq k^*) < c]^2.
        \end{aligned}
    \end{equation}
    Evaluated explicitly, we have
    \begin{equation}
    \label{eq:psyn}
        \begin{aligned}
        1 &- \epsilon(c; \sigma, p) = \\
        &\frac{1}{8} \left(\operatorname{erf}\left(\frac{c}{\sqrt{2 \sigma ^2-M (2 p (p+1)-1) \left(2 \sigma ^2+1\right)}}\right)+1\right)^2\times\\
        &\text{erfc}\Bigg\{\mkern-7mu\left(2^9 \pi ^3 c \sigma ^6-e^{-6 \pi ^2 \sigma ^2}M (2 (p-3) p+1) \operatorname{erf}\left(\sqrt{2} \pi  \sigma \right)^6\right)\\
        &\quad\quad\;\;\bigg/\bigg[-\bigg(2 M (p (3 p-7)+1) e^{-24 \pi ^2 \sigma ^2} \bigg(e^{12 \pi ^2 \sigma ^2}\times\\
        &\quad\quad\quad\quad\quad\quad\quad\operatorname{erf}\left(\sqrt{2} \pi  \sigma \right)^{12}-4 \pi ^3 \sigma ^6 \operatorname{erf}\left(2 \sqrt{2} \pi  \sigma \right)^6\bigg)\bigg)\\
        &- 2^{18} \pi ^6 \sigma^{12} \left(M (2 p (p+1)-1) \left(2 \sigma ^2+1\right)-2 \sigma ^2\right)\bigg]^{1/2}\Bigg\},
        \end{aligned}
    \end{equation}
    where the decoding step activation function cutoff \(c\) is obtained by maximizing the probability of success over all possible values of \(c\).
    This results in the logical error rate \(\epsilon(\sigma, p) := \min_c \epsilon(c; \sigma, p)\).

\end{document}